\documentclass[12pt]{iopart}

\usepackage{iopams}  

\usepackage{cite}  

\usepackage{tikz}

\usepackage{amsfonts}
\usepackage{bbm}

\usepackage{algorithm}      
\usepackage[noend]{algpseudocode}   
\algrenewcommand\alglinenumber[1]{\scriptsize #1:}

\usepackage{listings}       

\usepackage{bbold}

\usepackage{url}

\usepackage{subfig}
\usepackage{graphicx}

\usepackage{hyperref}
\usepackage{comment}

\usepackage{orcidlink}

\newcommand{\enc}{\mathrm{Enc}}
\newcommand{\dec}{\mathrm{Dec}}
\newcommand{\kmeans}{\mathrm{KMeans}}
\newcommand{\ikmeans}{\mathrm{IKMeans}}
\newcommand{\height}{\mathrm{height}}
\newcommand{\width}{\mathrm{width}}
\newcommand{\depth}{\mathrm{depth}}

\usepackage{pifont}
\newcommand{\cmark}{\ding{51}}%
\newcommand{\xmark}{\ding{55}}%

\definecolor{fiorentina}{RGB}{72, 46, 146}

\begin{document}

\def\aligned{\vcenter\bgroup\let\\\cr
\halign\bgroup&\hfil${}##{}$&${}##{}$\hfil\cr}
\def\endaligned{\crcr\egroup\egroup}

\title[Datacube segmentation via Deep Spectral Clustering]{Datacube segmentation via Deep Spectral Clustering}

\author{Alessandro Bombini\orcidlink{0000-0001-7225-3355}${}^{1,2}$, Fernando García-Avello Bofías\orcidlink{0000-0001-6640-8736}${}^{1}$, Caterina Bracci${}^{3,4}$, Michele Ginolfi\orcidlink{0000-0002-9122-1700}${}^{3,4}$, Chiara Ruberto\orcidlink{0000-0003-0321-7160}${}^{1,3}$.
}

\address{${}^{1}$ Istituto Nazionale di Fisica Nucleare (INFN), Via Bruno Rossi 1, 50019 Sesto Fiorentino (FI), Italy}
\address{${}^{2}$ ICSC - Centro Nazionale di Ricerca in High Performance Computing, Big Data \& Quantum Computing, Via Magnanelli 2, 40033 Casalecchio di Reno (BO), Italy}
\address{${}^{3}$ Dipartimento di Fisica, Università degli Studi di Firenze, Via Giovanni Sansone 1, 50019 Sesto Fiorentino (FI), Italy}
\address{${}^{4}$ Istituto Nazionale di Astrofisica (INAF), Osservatorio Astrofisico di Arcetri, Largo E. Fermi 5, 50125, Firenze (FI), Italy}

\ead{bombini@fi.infn.it}

\vspace{10pt}
\begin{indented}
\item[]\today
\end{indented}

\begin{abstract}
Extended Vision techniques are ubiquitous in physics. However, the data cubes steaming from such analysis often pose a challenge in their interpretation, due to the intrinsic difficulty in discerning the relevant information from the spectra composing the data cube. 
Furthermore, the huge dimensionality of data cube spectra poses a complex task in its statistical interpretation; nevertheless, this complexity contains a massive amount of statistical information that can be exploited in an unsupervised manner to outline some essential properties of the case study at hand, e.g.~it is possible to obtain an image segmentation via (deep) clustering of data-cube's spectra, performed in a suitably defined low-dimensional embedding space. 
To tackle this topic, we explore the possibility of applying unsupervised clustering methods in encoded space, i.e.~perform Deep Clustering on the spectral properties of datacube pixels. A statistical dimensional reduction is performed by an ad hoc trained (Variational) AutoEncoder, in charge of mapping spectra into lower dimensional metric spaces, while the clustering process is performed by a (learnable) iterative K-Means clustering algorithm. 
We apply this technique to two different use cases, of different physical origins: a set of Macro mapping X-Ray Fluorescence (MA-XRF) synthetic data on pictorial artworks, and a dataset of simulated astrophysical observations. 

%
\end{abstract}

%
\vspace{2pc}


\section{Introduction}

The advent of Machine Learning (ML), and more specifically Deep Learning, has revolutionised numerous fields, from computer vision to natural language processing \cite{Goodfellow-et-al-2016}. In recent years, these powerful computational tools have also begun to reshape the field of applied physics.

Deep Learning, a subset of ML, employs artificial neural networks with multiple layers (hence ``deep'') to model and understand complex patterns. In the realm of applied physics, Deep Learning can be used to recognise patterns in large datasets, predict system behaviour, or control physical systems.

One of the key advantages of Deep Learning in applied physics is its ability to process and learn from vast amounts of data, often surpassing traditional computational physics methods in both speed and accuracy. This is particularly useful in fields where experimental or simulation data is abundant, such as astrophysics or nuclear physics applied to Cultural Heritage.

In astrophysics, the growing influx of data from advanced astronomical facilities calls for the use of Deep Learning algorithms, which excel in modeling complex patterns and handling large-scale data, often outperforming traditional data analysis methods (see \cite{Huertas_Company_2023, Smith_2023}, and references therein).
In the research field of galaxy formation and evolution, Deep Learning methods, such as Convolutional Neural Networks (CNNs), have shown remarkable proficiency. Recent studies \cite{Huertas_Company_2015, Dieleman_2015, huertascompany2023galaxy} have employed these methods for more accurate and efficient galaxy morphological classification, a critical aspect in understanding galaxy evolution. Similarly, Deep Learning has revolutionized the study of gravitational lensing. For instance, the work \cite{Hezaveh_2017} illustrates how deep neural networks can analyze Hubble Space Telescope images to measure galaxy mass distribution more precisely, thus enhancing our understanding of cosmic structures.
The field of exoplanet discovery has also greatly benefited from Deep Learning (see e.g., \cite{Shallue_2018, Jin_2022}). As an example, \cite{Valizadegan_2022} have successfully applied neural networks to Kepler mission data, significantly improving the detection rate of new exoplanets. In cosmology, the analysis of the Cosmic Microwave Background has been revolutionized by Deep Learning (see, e.g., \cite{Wang_2022, Moriwaki_2023}. and references therein), as highlighted in recent works \cite{Caldeira_2019, adams2023cosmic}, which have extracted subtle cosmological signals from complex datasets more effectively than traditional methods. Moreover, the application of Deep Learning in transient astrophysical phenomena, such as fast radio bursts and supernovae, has enabled rapid data processing and analysis, crucial for timely scientific observations (e.g., \cite{Connor_2018, F_rster_2022}).
In summary, the growing field of Deep Learning in astrophysics is not just a technical advancement but a necessary evolution to keep pace with the ever-growing complexity and volume of astronomical data.

In the field of nuclear physics applied to Cultural Heritage (for a nice introduction to the subject, see, e.g.~\cite{alma991021183529704336, KnollBookXRF, MandoPixe, Grieken1993HandbookOX, jenkins1995quantitative, Janssens2000MicroscopicXF, Verma2007AtomicAN, Ruberto2023} and references therein), Deep Learning can be used to analyze data from non-destructive testing techniques, such as neutron imaging or gamma-ray spectroscopy. For example, it can help identify the composition of ancient artifacts or detect hidden structures in historical buildings, providing valuable insights without damaging these precious objects.
%
%
This involves the use of advanced algorithms to statistically analyse the data obtained from imaging using nuclear techniques. These algorithms can identify patterns and correlations in the data, providing valuable insights into the artwork’s composition, age, and condition. \cite{Kleynhans2020,5967899,8432512,8667664,666999xrfdl, D2JA00114D, Jones2022, arxiv.2207.12651, BombiniICIAP2021, BombiniICCSA2022, bombini2023ganx, BombiniThespianXRF, Dirks2022AutoEncoderNN, Liu2023NeuralNF} (for other Machine Learning approaches in Cultural Heritage, see \cite{FIORUCCI2020102}, and references therein). 

Out of many Deep Learning methods, Deep Clustering is one approach that may be relevant for the topic at hand (for nice reviews of Deep Clustering, see \cite{aljalbout2018clustering, 8412085, Nutakki2019, WEI2024127761, ren2022deep, zhou2022comprehensive}, and references therein). Deep Clustering is a branch of unsupervised learning that combines traditional clustering methods with Deep Learning.

The goal of classical clustering is to partition a dataset into groups, called \textit{clusters}, such that data points in the same cluster are more similar to each other than to those in other clusters. Traditional clustering methods, such as K-means \cite{kmeans1967originalpaper}, rely on predefined distance metrics to measure the similarity between data points. However, these methods often fail to capture complex patterns in high-dimensional data.

This is where Deep Clustering comes into play. Deep Clustering leverages the power of deep neural networks to learn a suitable data representation for clustering in an end-to-end fashion. The deep neural network is trained to map the input data into a lower-dimensional space where the traditional clustering algorithm is applied. The key idea is to learn a mapping that makes the clustering task easier in the transformed space.

There are several approaches to Deep Clustering, including autoencoder-based methods \cite{song2013auto, mrabah2020deep}, deep embedded clustering \cite{yang2017kmeansfriendly, guo2017improved}, and methods based on self-supervised learning \cite{yang2016joint, shah2018deep, van2020scan, gidaris2018unsupervised}. Autoencoder-based methods use an autoencoder \cite{10.5555/104279.104293} to learn a compressed representation of the input data, and then perform clustering on the compressed data. Deep embedded clustering extends this idea by jointly optimizing the autoencoder and the clustering objective.

In self-supervised learning methods, auxiliary tasks are designed to exploit the inherent structure of the data, and the learned representations are used for clustering. For example, a network might be trained to predict the rotation of an image, and the learned features are then used for clustering.

Deep Clustering has shown promising results in various applications. We explore here the possibility of using such method to perform datacube segmentation via deep spectral clustering.  

In Section \ref{sec:methods}, we describe the deep embedding models we will use. In Section \ref{sec:results}, we train deep embedding models in two use cases, using synthetic dataset, an astrophysical synthetic dataset, and a synthetic heritage scientific dataset. Finally, in Section \ref{sec:conclusions} we report the conclusions and the planned future steps.


\section{Methods: Deep Clustering with AutoEncoders } \label{sec:methods}

The core idea of this report is to obtain datacube segmentation by performing unsupervised clustering on the dimensionally reduced latent space. The whole process is self-supervised, since only spectra are used during training. The whole architecture comprises:
\begin{enumerate}
    \item An Encoder: $\enc : X \in \mathcal{M}\subseteq \mathbb{R}^{d=512} \mapsto \mu \in \mathcal{N}\subseteq \mathbb{R}^{n \ll 512}$, $\mu = \enc [X]$; 
    \item A clustering algorithm in latent space: $\ikmeans: \mu^{(i)} \mapsto C_{I=1, \ldots N}$
    \item A Decoder: $\dec : \mu \in \mathcal{N}\subseteq \mathbb{R}^{n \ll 512} \mapsto  X \in \mathcal{M}\subseteq \mathbb{R}^{d=512} $, $\tilde X = \dec [\mu]$;
\end{enumerate}

Having a well-trained architecture of that sort, starting from a datacube $\mathcal{I}_{x,y; e}$ of shape $(\width, \height, \depth)$, where the first two indices represent the $x-$ and $y-$pixel position, while $e$ is the spectral index, we can extract the $\height \cdot \width$ $\depth-$dimensional spectra $X^{(i=1,\ldots \height \cdot \width)}$, pass them to the Encoder to get a lower-dimensional statistical representation of the data $\mu^{(i)} = \enc[X^{(i)}]$, and perform a clustering onto those, to get a set of clusters $C_{I=1, \ldots N}$, and their barycenter $c_I = \frac{1}{\# C_I} \sum_{j | \mu^{(j)}\in C_I} \mu^{(j)} $. By using the Decoder it is now possible to see how the barycenter maps in the spectral space, $\bar{X}_I = \dec[c_I]$; furthermore, it is possible to create binary maps of clusters onto the image, by retaining the pixel-to-index relation, $(x,y) \leftrightarrow i$. We will show that those binary maps effectively represent unsupervised segmentation of the datacubes.

In the following, we are going to describe the implementation of the Deep Clustering architecture, based on the Deep Clustering Network (DCN) architecture of \cite{yang2017kmeansfriendly}.

\subsection{AutoEncoders Architecture}

\begin{figure}[ht]
    \centering
    \includegraphics[width=0.95\textwidth]{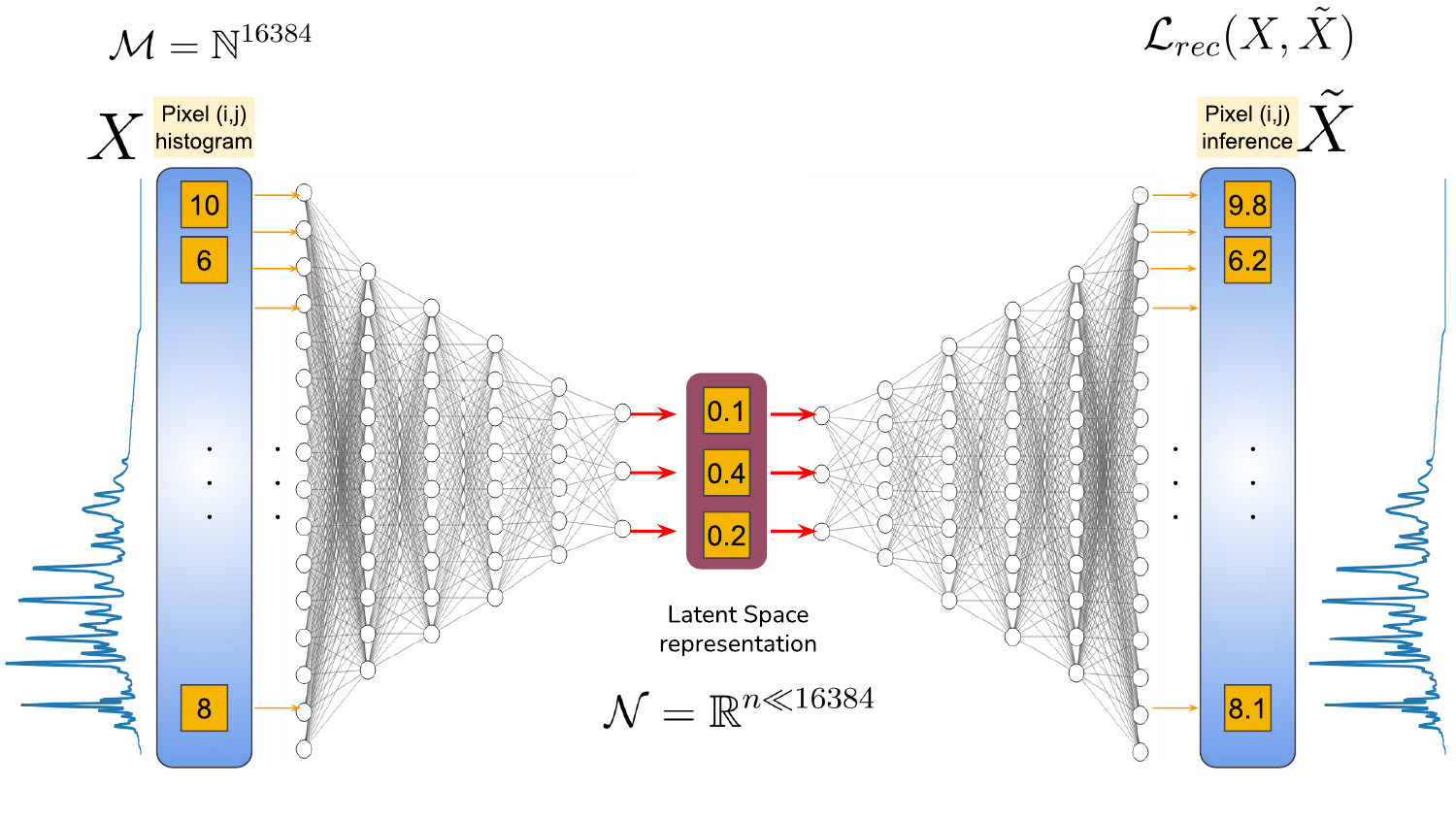}
    \caption{Graphical representation of an Autoencoder architecture.}
    \label{fig:AEarch}
\end{figure}

The first ingredient is an AutoEncoder (AE) architecture \cite{Goodfellow-et-al-2016, https://doi.org/10.1002/aic.690370209,  MAL-056, feickert2021living}, based on either on plain Multi-Layer Perceptrons (MLP) with Drop-out layers, or Self-Normalising Neural Networks (SNN) \cite{klambauer2017selfnormalizing}. The AutoEncoder class can be instantiated with a tunable number of layers, while the nodes are algorithmically fixed, starting from the Input size and dividing either by $2\cdot k$ (where $k$ is the layer index), or by $2^{k}$, up to the latent space dimension $n$. The Encoder and Decoder are instantiated with a specular architecture, i.e. the first Encoder layer has the same size as the last Decoder one, and so on inwards. The layers' weights are initialised using the Kaiming-He initialisation \cite{he2015delving}. For the MLP module, the layer's activation is a ReLU and the dropout probability can be tuned, while of course, the activation for the SNN module is a SELU. 
For a visual representation of an autoencoder, see figure \ref{fig:AEarch}.

The reconstruction loss used is the Mean-squared error, 
\begin{equation}
    \mathcal{L}_{rec} (X, \tilde X) = \frac{1}{N_{batch}} \sum_{a=1}^{N_{batch}} \frac{1}{d} \sum_{i=1}^d \left| X_{a, i} - \tilde X_{a, i} \right|^2 \,.
\end{equation}

\subsection{Iterative K-Means}

The clustering algorithm performed in the embedding space $\mathcal{N}$ is an iterative version of the standard K-Means clustering algorithm \cite{zbMATH03340881}, where the number of clusters $K$ is obtained by optimizing the \textit{silhouette score} \cite{ROUSSEEUW198753}.
The K-Means initialisation can be selected to either be random points or k++ \cite{kplusplus}, while the algorithm implementation is the standard Lloyd's algorithm \cite{LloydAlgo}.

The iterative selection works as explained in Algorithm \ref{alg:IKMeans}\footnote{For an overview of applications of unsupervised metric in K-Means clustering, see \cite{DEAMORIM2015126} and references therein.} (dubbed IKMeans).
\begin{algorithm}[H]
\caption{Iterative K-Means}\label{alg:IKMeans}
\begin{algorithmic}[1]

\Procedure{Iterative KMeans}{$X$, $N_i$, $N_f$}

\For{$k \in [N_i, N_{f}]$} \Comment{Iterate over $k$ clusters}
    \State $X^{(i)} \in C_{I=1, \ldots k} \leftarrow \kmeans[X]$; \Comment{Perform K-Means}
    \State $s_k \leftarrow \mathrm{silhouette}[C_I]$; \Comment{Compute silhouette score}
    
\EndFor 

\State Select $\hat{k} : \max_{k} s_k = s_{\hat{k}}$ \Comment{Pick $k$}

\Return $s_{\hat{k}}, C_{I=1, \ldots \hat{k}}$

\EndProcedure
\end{algorithmic}
\end{algorithm}

The Silhouette score is computed as \cite{silhouettescore1990}
\begin{equation}
 \begin{aligned}
     &\mathrm{silhouette}[C_{I=1, \ldots k}] = \frac{1}{k} \sum_{I=1}^{k} \frac{1}{\# C_I}  \sum_{i=1}^{\# C_I} \frac{b_I (i) - a_I(i) }{\max_i [a_I(i), b_I(i)]} \,, \\
        &a_I(i) = \frac{1}{\#C_I - 1} \sum_{j \in C_I, i\neq j} \mathrm{dist}(i,j) \,, \quad b_I(i) =  \min_{J \neq I}  \frac{1}{\#C_J } \sum_{j\in C_I} \mathrm{dist}(i,j) \,,
 \end{aligned}
\end{equation}
i.e.,~$a_I(i)$ is the mean intra-cluster distance, while $b_I(i)$ is the smallest mean distance of the point $i$ to all points in any other cluster (the extra-cluster distance). Notice that $\mathrm{silhouette}[C_{I=1, \ldots k}] \in [-1, +1]$, where values closer to $-1$ means very ill-formed clusters, while values closer to $+1$ means well formed clusters. 

\subsection{Deep Clustering}

Assembling the ingredients introduced in the sections above, the autoencoder architecture and the iterative K-Means clustering, it is possible to build a Deep Clustering autoencoder, by defining the total loss as
\begin{equation}
    \mathcal{L}_{tot} (X, \tilde X) = \mathcal{L}_{rec} (X, \tilde X) + \gamma \mathcal{L}_{sil} (X, \tilde X) \,,
\end{equation}
where we have defined
\begin{equation}
    \mathcal{L}_{sil} (X, \tilde X) \equiv \frac{1 - \mathrm{silhouette}[C_{I=1, \ldots k}]}{2} \,, \quad C_{I=1, \ldots k} \leftarrow \mathrm{IKMeans}[\enc[X]] \,.
\end{equation}

\begin{figure}[ht]
    \centering
    \includegraphics[width=0.95\textwidth]{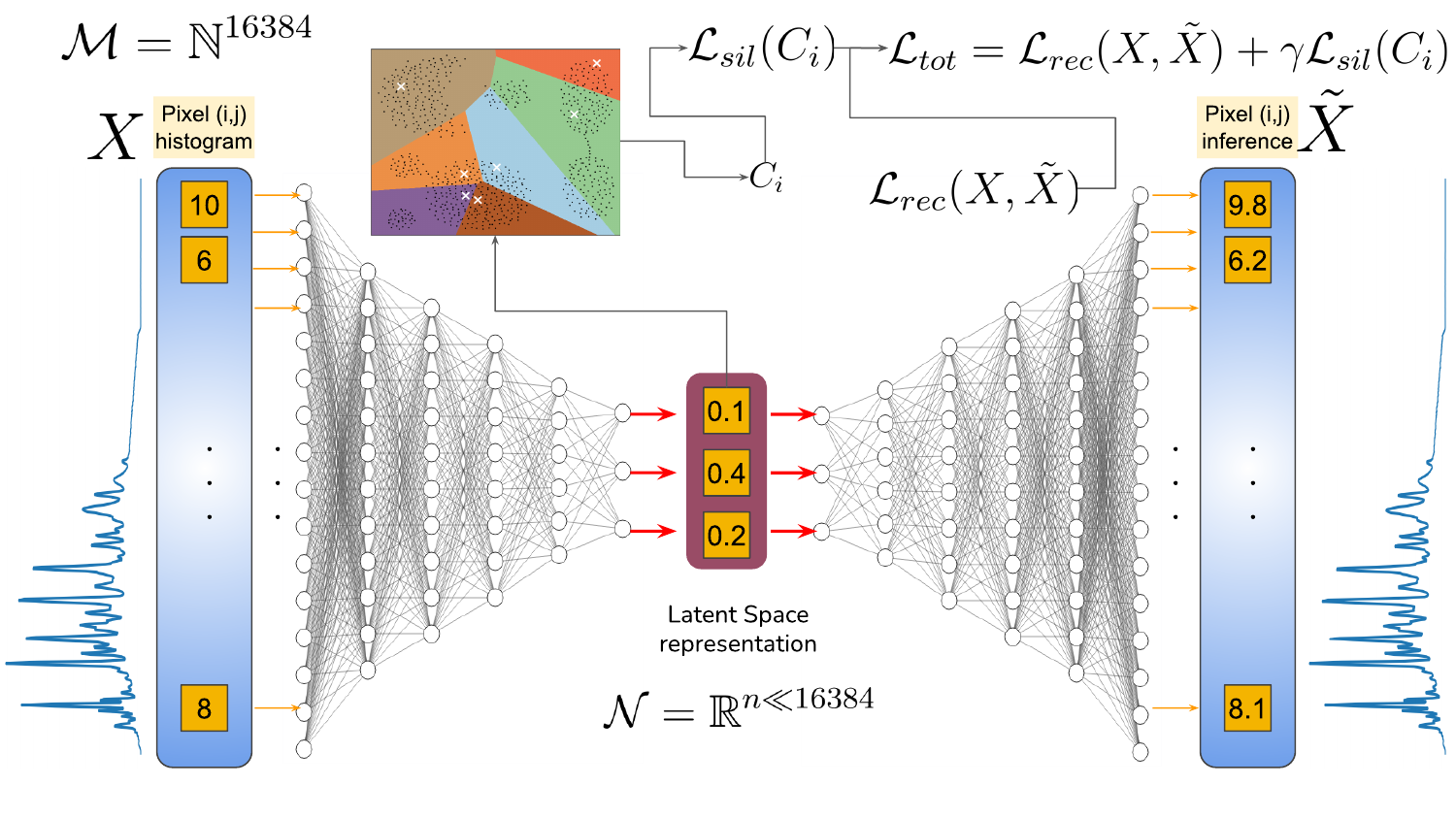}
    \caption{A visual representation of the Deep Clustering architecture. }
    \label{fig:DeepClusteringarchitecture}
\end{figure}

It is important to notice that the implementation of the IKMeans clustering algorithm has to be written in a way that it does not break the back-propagation algorithm. We implemented IKMeans as a \textsc{PyTorch} module, which ensures adherence to the differentiable programming paradigm. 

\subsection{A Deep Clustering Variational AutoEncoder? Variational Deep Embedding}

This approach suggests that we can move even further. We can use the approach of $\beta$-VAE \cite{Higgins2016betaVAELB} (for more details on Variational autoencoders, see \cite{Goodfellow-et-al-2016, Kingma_2019, BengioRepLernAE, prince2023understanding}, and references therein). The use of $\beta$-VAE has been explored in the context of Deep Clustering in a few seminal papers \cite{dilokthanakul2017deep, VAEDC11, Lim2020DeepCW}, giving rise to what is called \textit{Variational Deep Embedding} (VaDE) models \cite{jiang2017variational}. 

In addition, to address the well-known issues of \textit{posterior collapse} \cite{bowman2016generating, Alemi2017FixingAB, Lucas2019UnderstandingPC, lucas2019dont} and \textit{information preference} problem \cite{chen2017variational} during training, we employ an infoVAE architecture \cite{zhao2018infovae, Zhao2019InfoVAEBL}, i.e., instead of using the Kullback-Leibler divergence, we use the Maximum-Mean Discrepancy (MMD) approach \cite{gretton2008kernel, li2015generative}. MMD is a framework to quantify the distance between two distributions by comparing all of their
moments. We implemented it using the kernel trick: letting $k( \cdot, \cdot)$ be any positive definite kernel, the MMD between two distributions $p$ and $q$ is defined as 
\begin{equation}
    \mathrm{MMD} ( q || p) = \mathbf{E}_{p(z), p(z')} [k(z, z')] - 2  \mathbf{E}_{q(z), p(z')} [k(z, z')] + \mathbf{E}_{q(z), q(z')} [k(z, z')] .
\end{equation}

Additionally, we allow for either $\beta$ and/or $\gamma$ to be epoch-dependent, in order to dynamically tune how the MMD and the Silhouette terms affect the reconstruction during the training. The loss we use is
\begin{equation}
    \mathcal{L}_{tot}^{(i)} (X, \tilde X) = \mathcal{L}_{rec} (X, \tilde X) + \beta(i) \mathcal{L}_{\mathrm{MMD}} (\enc[X], \mathcal{N}(0, \mathbb{1}_d))    + \gamma (i) \mathcal{L}_{sil} (X, \tilde X),
\end{equation}
where $i$ is the $i-$th training epoch, and where $\mathcal{L}_{\mathrm{MMD}}  \equiv \mathrm{MMD} (\enc[X] || \mathcal{N}(0, \mathbb{1}_d) )$.

\subsection{Training vs Evaluation settings}

Before moving on into the topic of synthetic datasets generation, and how the Deep Embedding models are trained and thus evaluated, we 
discuss in depth the logical steps necessary to get from a Deep (Variational) Embedding of 1D spectra to a 3D datacube unsupervised segmentation algorithm. 

The model is trained on 1D spectra, i.e.~the pixel-wise content of a standard datacube. In fact, we refer to datacube as the data structure $I_{i,j;e}$ described by three indices, $\{i,j;e\}$; $(i,j)$ indices refer to the pixel position along $X$ and $Y$ image axes, while $e$ is the energy channel. In analogy, an RGB image is a datacube with three possible energy channels, corresponding to the Red, Blue and Green pixel content.  So, the RGB equivalent of the training process is to train a 1D model to embed single RGB triples into a different, low dimensional metric space; here, the spectra are multi-channels\footnote{They should actually be continuous, representing actual spectra (in the visible/near-infrared region for the Astrophysical case, in the X-ray region for the Cultural Heritage case); their discrete nature is of course given by the Analog-to-Digital converter which bins the data into discretised digital values.}, living in a very high dimensional space ($\mathbb{N}^{512}$ or $\mathbb{N}^{1024}$ once rebinned).

Notice that the model does not learn from images (i.e., it does not have access to \textit{geographical} information, like pixels proximity or location); it is trained over (synthetic) 1D signals.

The training of the Deep Neural Network model is done using mini-batch gradient descent (with \textsc{Adam} \cite{kingma2017adam} optimiser). The batch size is kept quite large ($512$ $1D$ spectra per batch), in order to have a meaningful per-batch clustering; for each cluster, an iterative KMeans clustering is performed, to find a per-batch optimum number of clusters, in an overall fixed range of possible cluster numbers (which are training hyperparameters). 

The fact that IKMeans is a distance-based clustering algorithm should force the Encoder model to map spectra into a well-behaved low dimensional space, which has to be metric (where the distance map inducing the metric is the same used in the IKMeans algorithm). 

Furthermore, for each hyperparameter set training trial, a \textsc{ReduceLROnPlateau} learning rate scheduler is used, in its standard \textsc{PyTorch} implementation. 
We notice that our training is unsupervised and is based solely on 1D signals (not images, at this level). 

After having trained the best model for the task, we try to assess the quality of the trained network. To do so, we generate a synthetic datacube, starting from a seed RGB image (how this is done will be explained in more depth in the following sections); we then flatten the image (thus seeing it as a collection of spectra), pass it to the trained model, get the embedded version (by reverse the flattening) and related clustering. We will show in the following that, as an emergent phenomenon, we experience a sort of semantic segmentation of the datacube in this process.

\subsubsection{Alternative approaches: Deep Embedding vs Dimensional Reduction:} \label{subsubsec:DimRed}
The emergence of a semantic segmentation through clustering in lower dimensional embedded space can also be obtained using other Dimensional Reduction algorithms, such as \textit{Principal Component Analysis} (PCA) \cite{FRS1901LIIIOL, Hotelling1933AnalysisOA} (for a modern introduction to the subject, see \cite{JolliffePCA2002} and references therein) or Manifold Learning algorithms, such as \textit{t-distributed stochastic neighbour embedding} (t-SNE) \cite{NIPS2002_6150ccc6, JMLR:v9:vandermaaten08a} or \textit{Uniform Manifold Approximation and Projection} (UMAP) \cite{mcinnes2020umap} (for a recent review on manifold learning and non-linear dimensional reduction, see \cite{doi.org/10.1002/wics.1222, annurev:/content/journals/10.1146/annurev-statistics-040522-115238, pml1Book, melaskyriazi2020mathematical, wang2021understanding} and references therein). 

While applying dimensional reduction algorithms to a spectral datacube, followed by a clustering algorithm is a viable method to get an emergent datacube semantic segmentation, this approach is rather different from the one we propose here.

Firstly, a Deep Embedding model, having a Decoder network, is invertible. While PCA is itself invertible, being a linear mapping, Manifold Learning such as t-SNE and UMAP are inherently difficult to invert. 

Furthermore, t-SNE and UMAP (as well as other non-linear dimensional reduction algorithms) have a computational cost that scales rapidly with vector dimensions; this means that, usually, we need to perform a preliminary PCA reduction to an intermediate dimensional space \cite{JMLR:v9:vandermaaten08a}\footnote{In \cite{mcinnes2020umap}, instead of PCA, a preliminary Spectral embedding is performed by considering the 1-skeleton of the global fuzzy topological representation as a weighted graph. They show that, empirically, UMAP is faster than t-SNE while scaling with ambient dimension; still, the computational cost does increase hugely, and non-linearly, even though more slowly.}. Thus, while Deep Embedding models need a slow train phase, which may be slower compared to the other methods, it is fast in its inference phase.

Additionally, the embedding space may not have a metric meaning for non-linear dimensional reduction; this latter point may be relevant, because it means that distances in the embedded, low dimensional space are not proxies for similarity (i.e., nearby points are not necessarily more similar, nor far away points are necessarily more different).

PCA, as well as other linear dimensional reduction methods, are incapable of capturing complex, non-linear relations among data points, due to their linear nature. This also implies that PCA-reduced datasets may be difficult to cluster properly. 

Finally, the crucial difference is that, while the deep embedding model is trained on a dataset formed as a collection of 1D spectra, and then applied onto datacubes (seen as a collection of spectra) at inference, the other models do need to be trained for each image (to have a well tailored behaviour). 

In this sense, the two approaches are rather different:
\begin{enumerate}
    \item in Deep Clustering, we train on an ad-hoc dataset and apply it on datacubes, and see what happens.
    \item In dimensional reduction + clustering, we fit-and-predict on per-image basis and see what happens.
\end{enumerate}
The two methods are different and, in some cases, may be complementary, since they fulfil different goals. 

Finally, if we train a Deep Variational Embedding model, we add a \textit{generative} spin, which cannot, in any case, be obtained in the Dimensional Reduction case. 

In Table \ref{tab:DEvsPCAvsManLe} we summarise these differences.

\begin{table}[ht]
\centering
\begin{tabular}{c|ccc|}
                                & Deep Embedding & PCA                      & Manifold Learning        \\ \hline
Invertible                      & {\color[HTML]{00FE00} \cmark}    & {\color[HTML]{00FE00} \cmark} & {\color[HTML]{FE0000} \xmark} \\
Pure high dimensionality origin & {\color[HTML]{00FE00} \cmark} & {\color[HTML]{00FE00} \cmark} & {\color[HTML]{FE0000} \xmark} \\
Fast inference                  & {\color[HTML]{00FE00} \cmark} & {\color[HTML]{00FE00} \cmark} & {\color[HTML]{FE0000} \xmark} \\
Fast training                  & {\color[HTML]{FE0000} \xmark} & {\color[HTML]{00FE00} \cmark} &  {\color[HTML]{00FE00} \cmark} / {\color[HTML]{FE0000} \xmark} \\
Metric embedding space                  & {\color[HTML]{00FE00} \cmark} & {\color[HTML]{00FE00} \cmark} & {\color[HTML]{FE0000} \xmark} \\
Captures non-linear relations                  & {\color[HTML]{00FE00} \cmark} & {\color[HTML]{FE0000} \xmark}  & {\color[HTML]{00FE00} \cmark} \\
Clustering-apt                & {\color[HTML]{00FE00} \cmark} & {\color[HTML]{FE0000} \xmark}  & {\color[HTML]{00FE00} \cmark} \\
Image independent               & {\color[HTML]{00FE00} \cmark} & {\color[HTML]{FE0000} \xmark} & {\color[HTML]{FE0000} \xmark} \\
Generative               & {\color[HTML]{00FE00} \cmark} / {\color[HTML]{FE0000} \xmark} & {\color[HTML]{FE0000} \xmark} & {\color[HTML]{FE0000} \xmark}\\ 
\hline
\end{tabular}
\caption{Summary of differences between Deep Embedding, PCA and Manifold Learning algorithms.}
\label{tab:DEvsPCAvsManLe} 
\end{table}


\section{Results: Deep Clustering synthetic datacubes }\label{sec:results}

Since the work is an exploratory phase, we applied the developed architecture on two synthetic datasets (ordered alphabetically). The first is a synthetic dataset of astrophysical interest; 
the second, is a synthetic dataset of imaging obtained by nuclear techniques on pictorial artworks.

The detailed description of how each dataset is created is described in the relative section. 
In both cases, to generate the datacube we have used \texttt{ganX}, an open-source  \textsc{Python} package to create spectral datacubes starting from RGB images and a well-defined dictionary of rgb-spectral signal relations \cite{bombini2023ganx, Bombini_ganX_-_generate_2023}. 

All the datasets, trained models, and produced code is publicly available online. See Section \ref{sec:data-and-code} for more details.

\subsection{Deep Clustering Astrophysical synthetic datacubes} \label{subsec:Astro-Res}



To create a realistic synthetic dataset of datacubes with astrophysical meaning, we started from synthetic 1D signals belonging to three classes: \textit{HII} regions, \textit{planetary nebulae} and \textit{shock regions}. 
These classes represent three types of ionized gas nebulae found in galaxies, characterized by distinctive morphological features and ionization mechanisms. 
As a consequence, when observed, they produce optical spectra, which differ mainly in the relative intensities of their spectral lines.
Therefore, traditional classification diagnostics are generally based on the differences in specific line ratios but do not exploit the information contained in their entire spectra, e.g., the Baldwin–Phillips–Terlevich (BPT) diagrams \cite{Baldwin_1981}. These techniques generally suffer from theoretical limits (see e.g. \cite{10.1093/mnras/stz2594}), and often require human aid or a specific and complex analysis, thus they are less suitable for handling large datasets, obtained from current instruments with integral field units. This is why Machine Learning algorithms, capable of encoding the information from a wide sample of emission-line regions in the full spectra, are a valid choice.

We generate a collection of synthetic nebular optical spectra, based on photoionization models run with \texttt{CLOUDY} \cite{Ferland2013THE2R}, withdrawn from the Mexican Million Models database (3MdB) \cite{Morisset2014AVO}. 
We augment the number of models by combining them linearly using random coefficients. 
To create mock spectra starting from these models, we first select relevant emission lines to simulate, based on real data. In this case, we create spectra based on observations of the local galaxy M33 obtained with the Multi Unit Spectroscopic Explorer (MUSE)  \cite{Bacon_2010}. Hence, the selected emission lines are defined in the MUSE wavelength domain.
Simulated emission lines are modeled with Gaussian profiles, whose fluxes are given by the model line intensities, and whose widths consist of two components: an instrumental width, caused by the finite spectral resolution of the telescope, and a physical one, given by the velocity dispersion in the nebulae, where the latter is chosen randomly from a range of possible values. 
Then, Gaussian noise (with standard deviations based on real data) is added to the spectra.
Thus, for each model resulting from the linear combination of two of the ``original'' database models, we can create multiple spectra with different noise levels and line widths.
This process is based on solid, realistic physical assumptions and includes typical noise distributions and specific instrumental effects, ensuring the resulting spectra are highly realistic. 

\begin{figure}[t]
    \centering
    \includegraphics[width=0.25\textwidth]{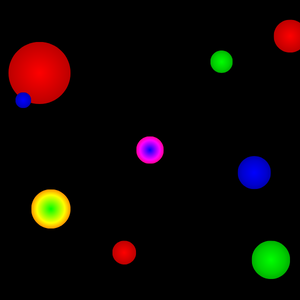}
    \caption{RGB seed image for the synthetic test datacube.}
    \label{fig:AstroDataCube}
\end{figure}

Then, we used \texttt{ganX} to create a datacube, using each synthetic signal, and assigning it (arbitrarily) to a colour; the mapping is the following: pure red ($[255,0,0]$) for the HII regions, pure green ($[0, 255,0]$) for the \textit{planetary nebulae} regions, pure blue ($[0,0,255]$) for the \textit{shock} regions. The RGB image used as a seed to generate the test datacube is reported in Figure \ref{fig:AstroDataCube}.

We trained a pure autoencoder model on $237.003$ synthetic multi-class spectra, divided into $161.001$ for the training set, $46.001$ for the validation set, and $30.001$ for the test set. We limited the size of the dataset to overcome the time constraint put on top of hardware limitations.

The model hyperparameters are obtained using a plain grid-search method. The Encoder has 3+1 layers, while the Decoder has 3 layers; the latent space has dimension $3$, the MLPs are Self-normalising Neural Networks \cite{klambauer2017selfnormalizing}, and the sizes are $1024\mapsto 512 \mapsto 256\mapsto  32\mapsto 3$ for the encoder, and $3 \mapsto  128 \mapsto 256 \mapsto  1024$ for the decoder. 

\subsubsection{Results:}

We applied the trained model on the synthetic datacube obtained using \texttt{ganX} \cite{bombini2023ganx} on the RGB data of Figure \ref{fig:AstroDataCube} with threshold $0.2$, giving rise to 4 clusters. 

\begin{figure}[ht]
    \centering
    \includegraphics[width=0.99\textwidth]{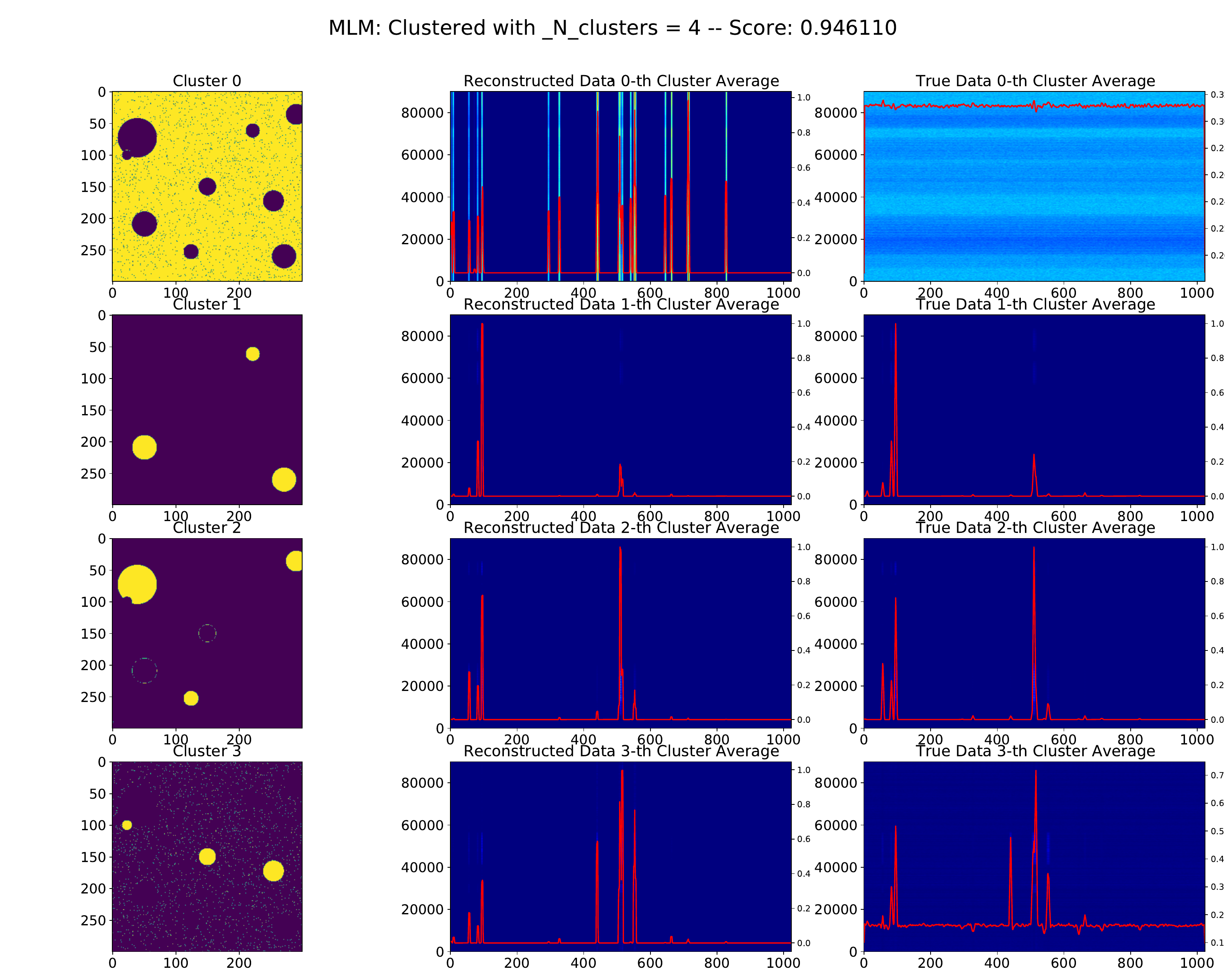}
    \caption{Clustered datacube using the trained model. On the left, is the binary map of the cluster pixels. In the middle, the Reconstructed Signal averaged over the cluster; on the right, the ``true" signal averaged over the cluster.}
    \label{fig:clustered_astro}
\end{figure}

The results are reported in Figure \ref{fig:clustered_astro}; the figure reads: on the left, the binary map of the cluster pixels, where pixels are colored in yellow if they belong to the cluster, and in purple if not. In the middle, the Reconstructed signal averaged over the cluster. Furthermore, the silhouette score obtained on the test datacube is $0.946110$, a very high result, due to the simplicity of the synthetic datacube. 

It is easy to recognise in Cluster 1 the nebulae region signals, in Cluster 2 the HII signals, in Cluster 3 the shock regions, and in Cluster 0 the noisy background; since no noise signal has been fed to the network during training, the decoder was not able to reproduce the signal; nevertheless, the clustering process in encoded space was able to isolate those signals, as expected. 

\begin{figure}[ht]
    \centering
    \includegraphics[width=0.95\textwidth]{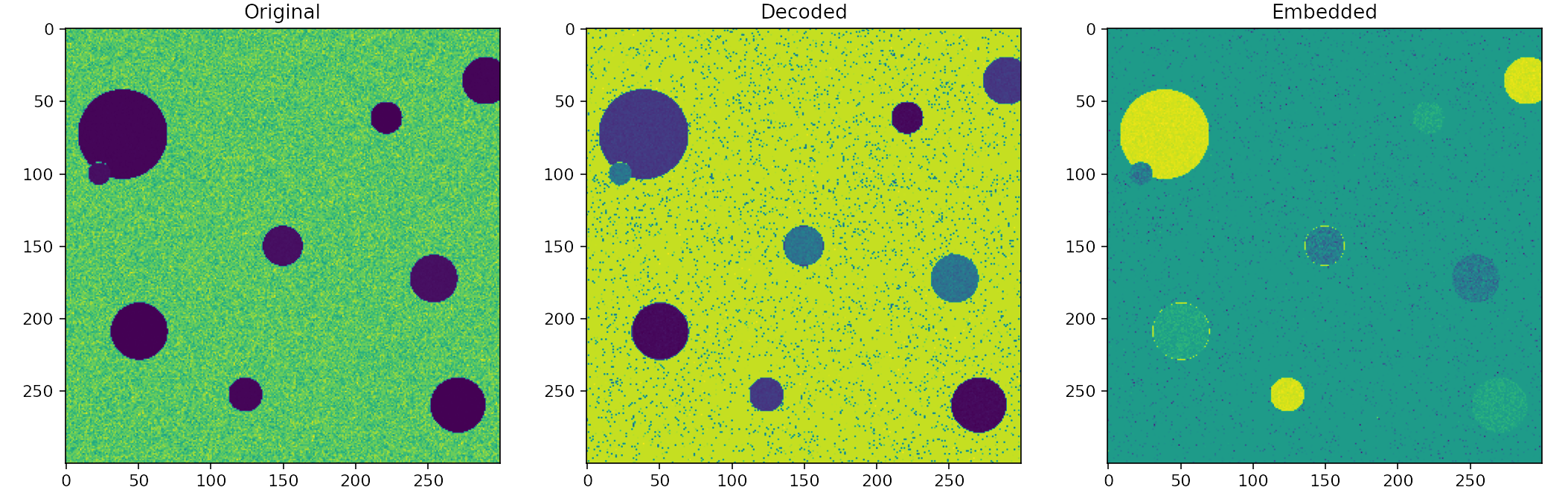}
    \caption{Energy integrated spectral data-cubes. }
    \label{fig:integrated_astro}
\end{figure}

Another relevant plot is Figure \ref{fig:integrated_astro}, which shows the energy-integrated data-cube, i.e. the gray-scale image (in \textit{viridis} color scale) obtained integrating over the whole energy channel. The left image in Figure \ref{fig:integrated_astro} represents the original, ``true"\footnote{Here, we refer to the ``true" signal as the input signal of the neural network, while with ``reconstructed" we refer to the output of the network.} spectral energy-integrated datacube $I_{i,j} = \sum_e I_{i,j ; e}$. The middle image is the decoded energy-integrated datacube $D_{i,j} = \sum_e \dec[\enc[I_{i,j ; e}]]$. It immediately shows noise reduction (or signal enhancement). The right image in Figure \ref{fig:integrated_astro} represent the embedded energy-integrated datacube $E_{i,j} = \sum_n \enc[I_{i,j ; n}]$, which further refine the readability.

\subsubsection{Check against RGB clustering:} \label{subsubsec:Astro_check_vs_rgb}

Since, as mentioned in Section \ref{sec:methods}, the datacube segmentation is an emergent result from clustering in latent space, and comes from a completely unsupervised approach, there is no direct evaluation check with standard semantic segmentation loss and measures, such as the Tversky Index (see, e.g., \cite{Jadon_2020,azad2023loss, 10.1007/978-3-031-51023-6_16}, and references therein).

Nervetheless, since we are discussing the application of synthetic datacubes, obtained starting from an RGB image (Figure \ref{fig:AstroDataCube}), we can perform the same Iterative K-Means clustering approach to the RGB image, and use it as a baseline to compare it to the clustering obtained with the model. 

This is a check that cannot be done in a real scenario, since 
\begin{enumerate}
    \item we may not have an RGB image;
    \item RGB clustering may not be able to capture the relevant information;
\end{enumerate}
Nevertheless, since our synthetic datacube generation pipeline relies on information extracted from the RGB image via a clustering process, we may use it as a proxy for a benchmark. 

\begin{figure}[t]
    \centering
    \includegraphics[width=0.9\textwidth]{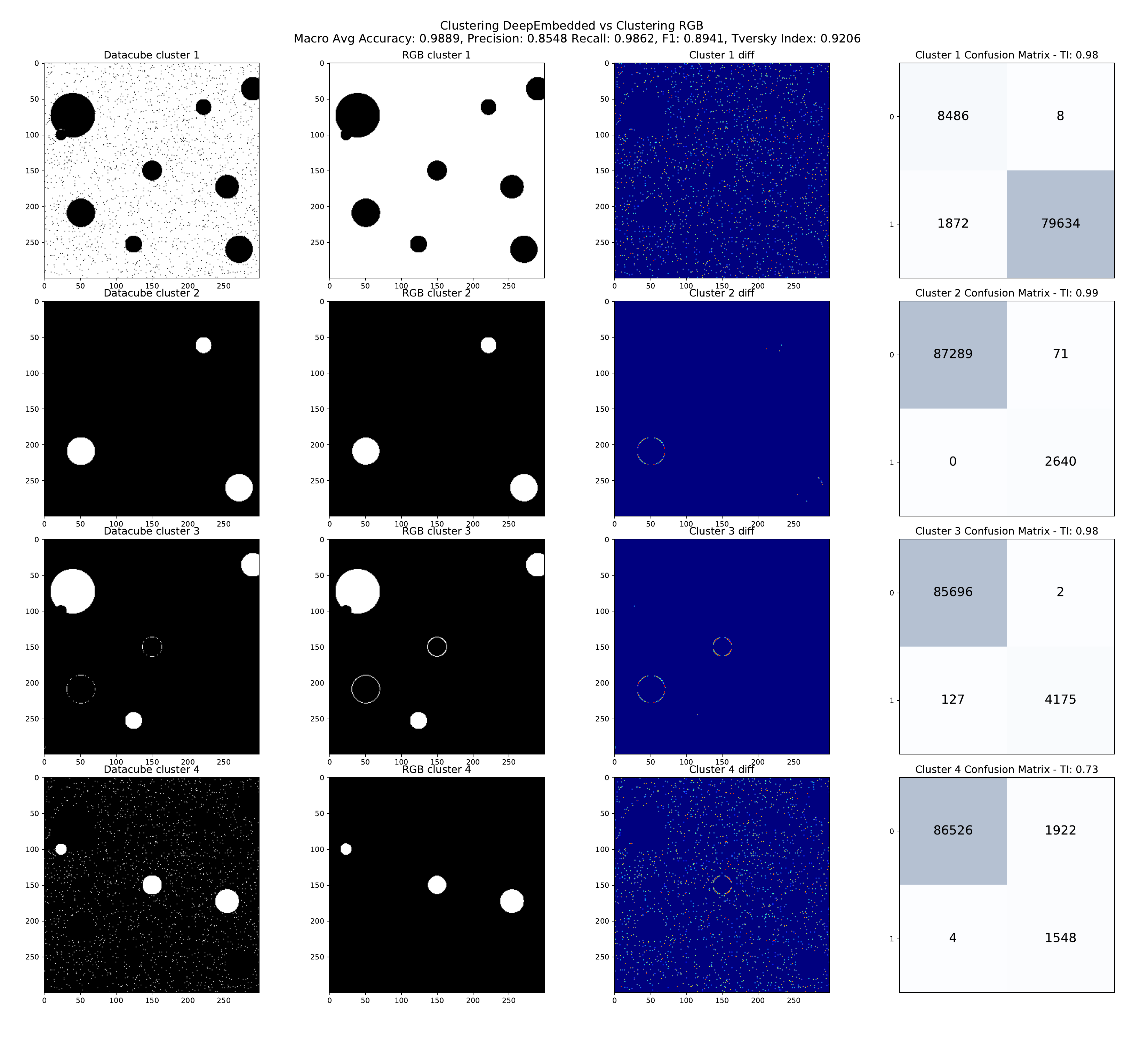}
    \caption{Deep Clustering Segmentation vs RGB Clustering Deep  Embedding model on Astrophysical synthetic datacube.}
    \label{fig:Astro_DeepEmbedVsRGB_clustering}
\end{figure}

In Figure \ref{fig:Astro_DeepEmbedVsRGB_clustering} we report the result of this check, done cluster-by-cluster. The columns are, from left to right: the black-and-white image of the cluster mask (i.e., pixels are white if they do belong to the cluster, and black if they don't) as obtained from the Deep Embedding Model; the same, but obtained from the RGB clustering; the image of mislabelled pixels (i.e., pixels appears red if they are mislabelled and blue if they are correctly labelled); the confusion matrix. On the confusion matrix is reported the number of counts (from left to right, from top to bottom: true positives, false positives, false negatives, true negatives). In the confusion matrix image subtitle is reported the computed Tversky Index. The Tversky Index (TI) formula is
\begin{equation}
    \mathrm{TI} = \frac{ \mathrm{tp} }{\mathrm{tp} + \alpha \, \mathrm{fn} + \beta \, \mathrm{fp}} \,, 
\end{equation}
we use the parameter set $\alpha = 0.7, \beta = 0.3$ as in \cite{abraham2018novel}.

Furthermore, in the Figure subtitle we reported the macro averages for Accuracy, Precision, Recall, F1 score and Tversky Index.

\begin{figure}
    \centering
    \includegraphics[width=0.6\textwidth]{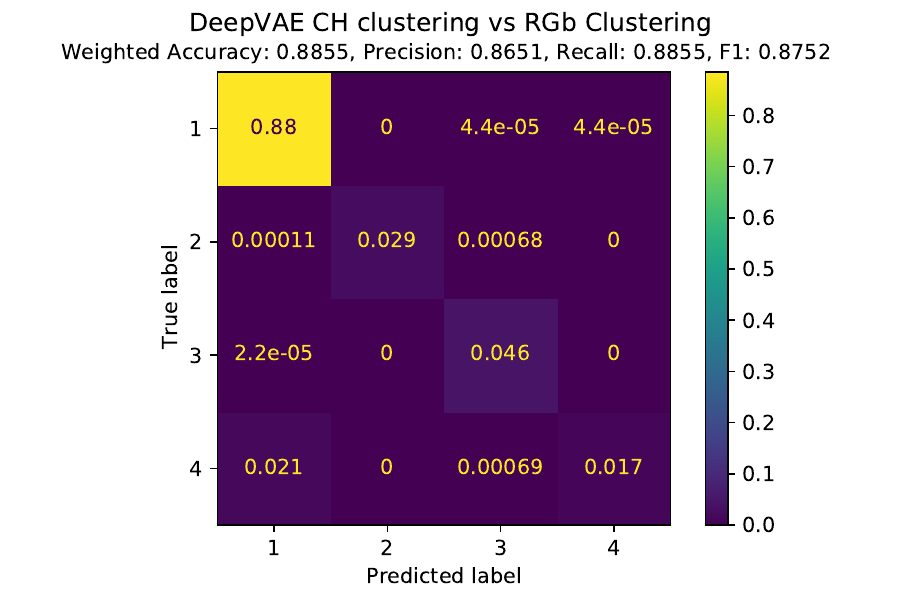}
    \caption{Multi-class Confusion Matrix of Deep Clustering Segmentation vs RGB Clustering on the synthetic Astrophysical Datacube.}
    \label{fig:Astro_MC-ConfusionMatrix}
\end{figure}

In Figure \ref{fig:Astro_MC-ConfusionMatrix} we report the multi-class confusion matrix; on each element of the confusion matrix is reported the counts normalised to the whole population. In the Figure subtitle is reported the Weighted average of Accuracy, Recall, Precision, and F1 score. 

Notice that the result obtained from the datacube has a large count in the square (true label, predicted label) $= (4,1)$; this, again, is due to the fact that the network has not seen any synthetic background signal during training, and yet it was still capable of cluster (most of) the background pixels. 

In \ref{app:noise} we report a test to show how the trained model behaves under the degradation of the signal (i.e., by adding noise such that the Peak-signal-to-Noise is reduced).


\subsection{Deep Variational Embedding of MA-XRF datacubes on Cultural Heritage} \label{subsec:CH-Res}

To create the synthetic dataset of imaging with nuclear techniques applied to Cultural Heritage assets, we employed the XRF pigment information from \cite{infraart10.1145/3593427}\footnote{The URL of the database is \url{https://infraart.inoe.ro/}.}; we have used the pigment palette of \cite{heritage2020103}, extended with few additional pigments that allow covering the set of most common Italian medieval choir books pigments \cite{SimiMasterThesis}. This allows us to create seven synthetic palettes (intended as different ensembles of pigment signals and associate RGB colour) with the nearest possible historical coherence within the synthetic dataset.

The RGB images are scraped from the \textit{Web Gallery of Art} \cite{WebGalleryOfArt}. In particular, we will show the application of the trained DNN on a test sample (never seen by the network during training) which is a miniature representing the \textit{Portrait of an Old Humanist}. It is a miniature contained in the \textit{Codex Heroica} by Phoilostratus, illuminated by the Florentine painter Attavante degli Attavanti (1487 - 1490 circa)\footnote{\url{https://www.wga.hu/frames-e.html?/html/a/attavant/heroica2.html}.}. The original is held at the National Széchényi Library, Budapest. 

\begin{figure}[ht] 
    \centering
    \subfloat[RGB of the test sample we use for analysis in the text. ]{%
        \includegraphics[width=0.35\textwidth]{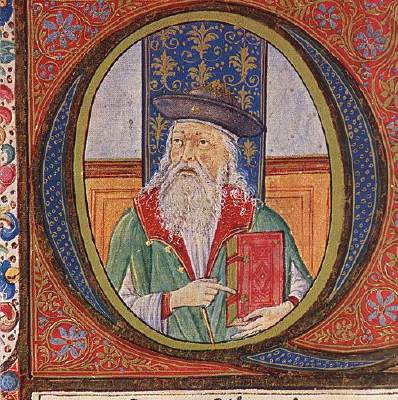}%
        \label{fig:1648_rgb}%
        }%
    \hfill%
    \subfloat[Clustered RGB obtained via iterative KMeans on RGB space. The found optimal K in the $(3,10)$ range was $4$.]{%
        \includegraphics[width=0.35\textwidth]{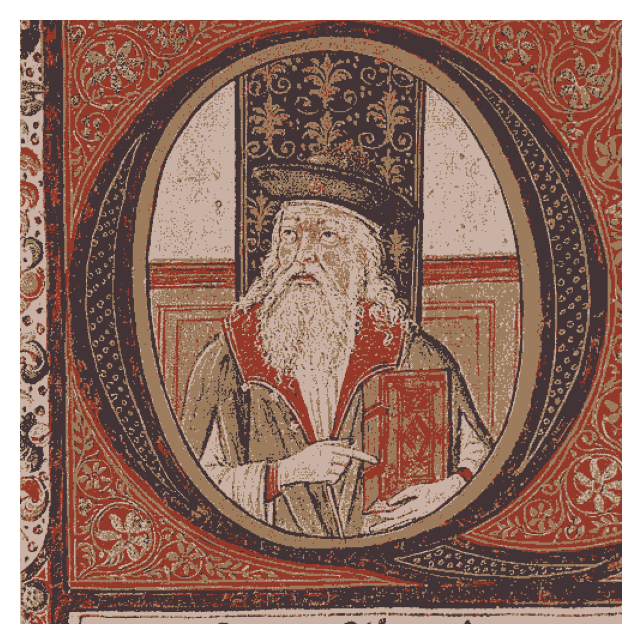}%
        \label{fig:1648_clustered_rgb}%
        }%
    \caption{Original RGB and Clustered RGB of the test sample used in the text. It is a miniature painted by the Florentine painter Attavante degli Attavanti and contained in the Codex Heroica by Phoilostratus (1487 - 1490 circa).}
\end{figure}

We trained the model on $891.276$ synthetic spectra, coming from 17 RGB miniatures, divided into $576.708$ for the training set, $209.712$ for the validation set, and $104.856$ for the test set. We limited the size of the dataset to overcome the time constraint put on top of hardware limitations. We plan to move to the whole synthetic dataset we built out of 1636 RGB images (around $10^8$ spectra).

The model hyperparameters are obtained using a plain grid-search method. Encoder has 4+1 layer, while the decoder has 4 layers; the latent space has dimension $3$, the MLP are Self-normalising Neural Networks \cite{klambauer2017selfnormalizing}, and the sizes are $512\mapsto 256\mapsto  128\mapsto  64\mapsto  32\mapsto 3$ for the encoder, and $3 \mapsto  64 \mapsto 128 \mapsto  256 \mapsto 512$ for the decoder. 

Furthermore, $\gamma(i) = 0.01$ $\forall i$, while $\beta(i) = 0.01 \cdot \Theta(i - 30)$. 

\subsubsection{Results:}

We applied the trained model on the synthetic XRF datacube obtained using \texttt{ganX} \cite{bombini2023ganx} on the RGB data of Figure \ref{fig:1648_rgb} with threshold $0.6$, which gave rise to 4 RGB clusters (Figure \ref{fig:1648_clustered_rgb})\footnote{We recall here the non-Riemannian nature of colour spaces \cite{doi:10.1073/pnas.2119753119}, which give rise to impossibilities of using metric approaches in colour space, leaving only distance-based methods.}.

\begin{figure}
    \centering
    \includegraphics[width=0.99\textwidth]{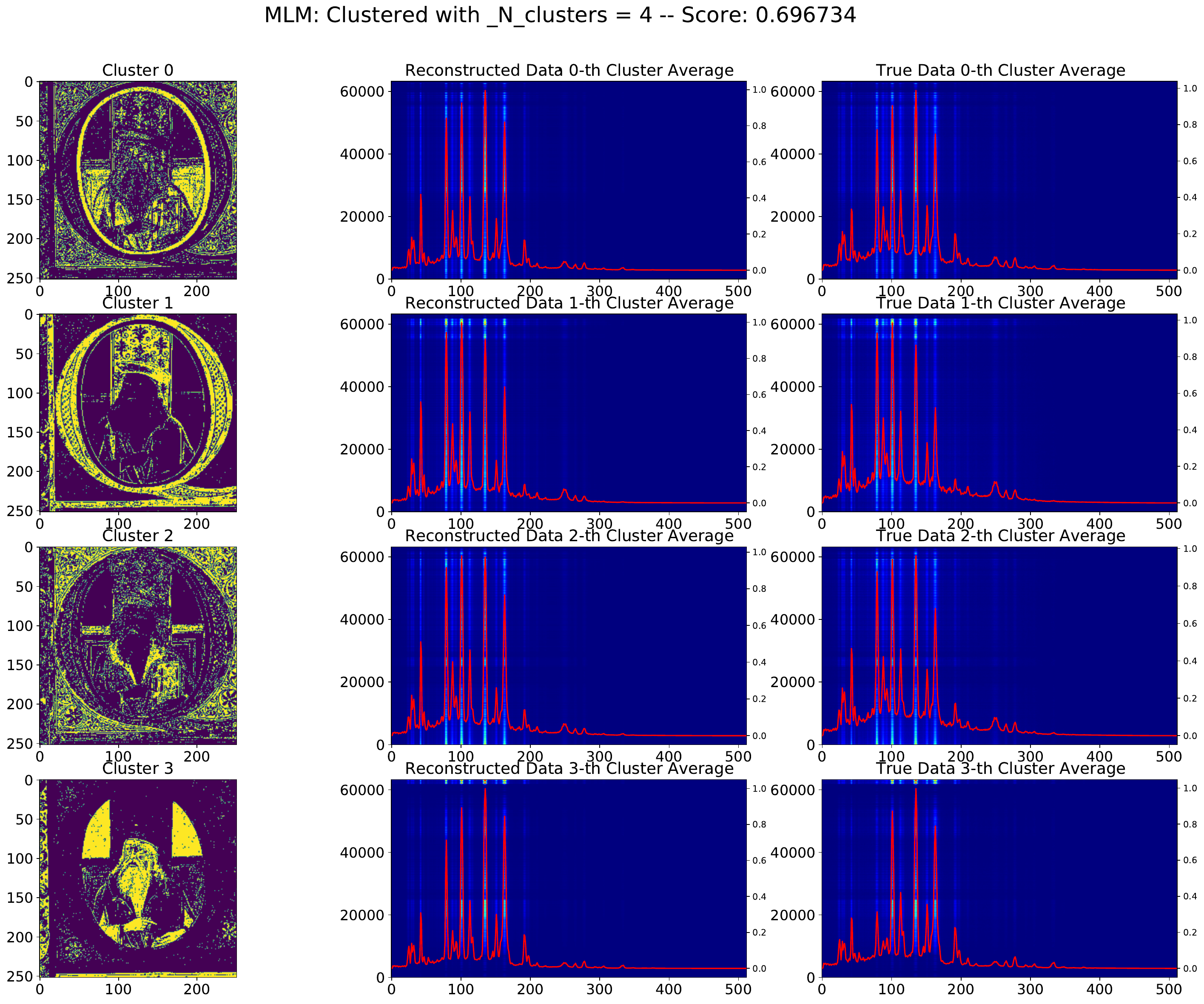}
    \caption{Clustered XRF using the trained Variational Deep Embedding trained model. On the left, the binary map of the cluster pixels. In the middle, the Reconstructed XRF averaged over the cluster; on the right, the ``true" XRF averaged over the cluster. }
    \label{fig:clustered_xrf}
\end{figure}

The results are reported in Figure \ref{fig:clustered_xrf}; the figure reads: on the left, the binary map of the cluster pixels, where pixels are coloured by yellow if they belong to the cluster, and by purple if not. In the middle, the Reconstructed XRF averaged over the cluster is reported, i.e.
\begin{equation}
    \bar{h}_I = \frac{1}{\# C_I} \sum_{a=1}^{\# C_I} \dec[\mu_a \in C_I] \,,
\end{equation}
where $\mu_a = \enc[X_a]$.

\begin{figure}[ht]
    \centering
    \includegraphics[width=0.99\textwidth]{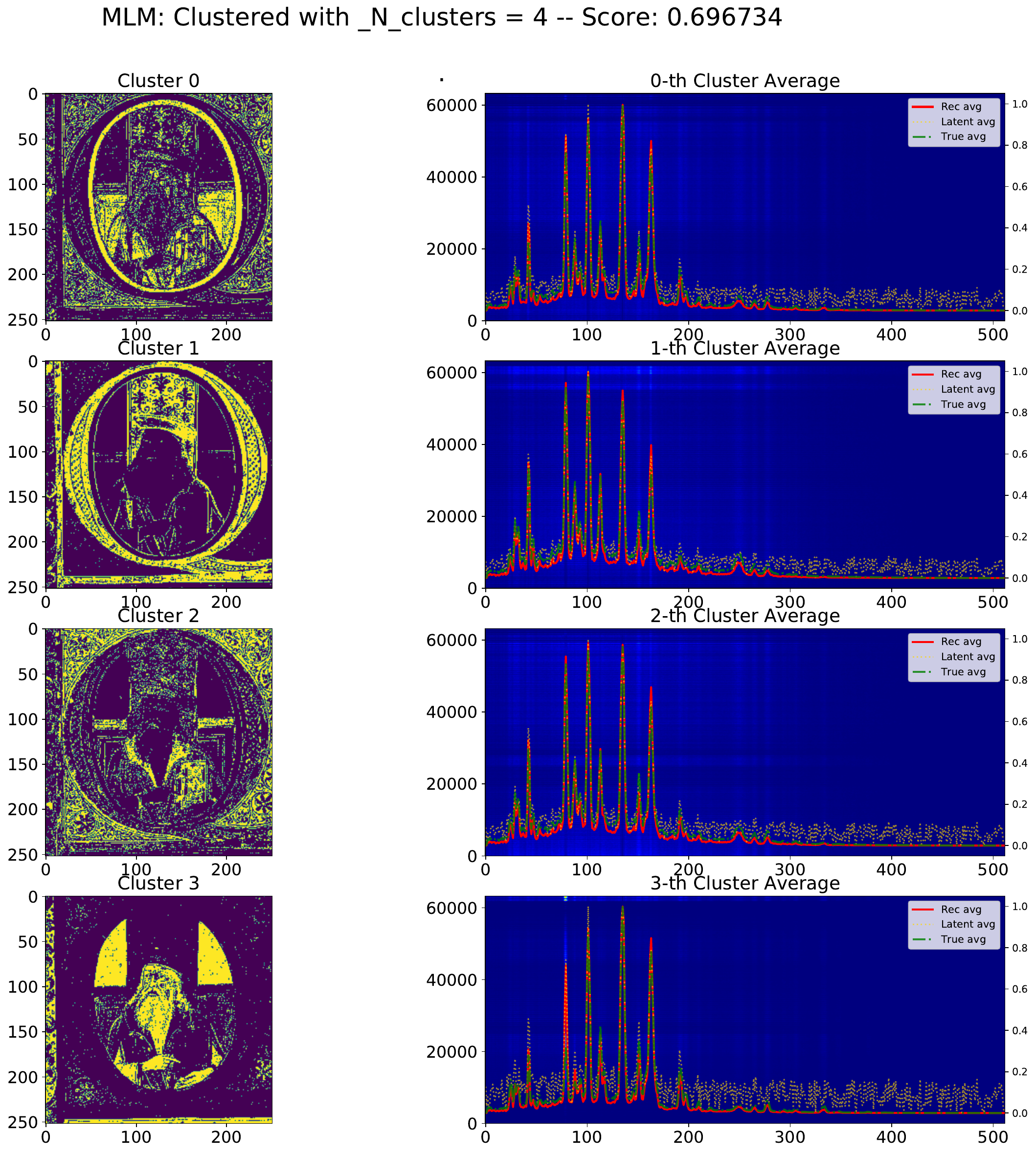}
    \caption{Plot reporting the Reconstructed, Latent and ``true" average plotted against each other in the same plot.}
    \label{fig:clustered_histograms}
\end{figure}

Similarly, on the right for Figure \ref{fig:clustered_xrf}, and in dash-dotted green for Figure \ref{fig:clustered_histograms}, the ``true" XRF averaged over the cluster, as well as the colormap of the flattened ``true" XRF\footnote{Here we use the quote marks on the word \textit{true}, because the XRF is synthetic. Nevertheless, it is the ``true" XRF in the sense it is the input from the DNN architecture perspective.  }, i.e.:
\begin{equation}
    \bar{h}_I^{(true)} = \frac{1}{\# C_I}  \sum_{a=1}^{\# C_I} X_a
\end{equation}

Figure \ref{fig:clustered_histograms} shows additional plots: in solid red, the aforementioned average over the cluster in the reconstructed space; in dashed gold, there is the decoded average in latent space, i.e.
\begin{equation}
    \langle {h}_I \rangle = \dec\left[ \frac{1}{\# C_I}  \sum_{a=1}^{\# C_I} \mu_a \in C_I \right] \,.
\end{equation}

For a geometric intuition, the former is the center-of-mass of the reconstructed cluster, seen as a set of vectors in the high-dimensional original space; the latter, instead, is the reconstruction (through the decoder) of the center-of-mass of the cluster computed in the low-dimensional latent space. 

Evidently, the two do not match, due to the non-linearity of the decoder. Nevertheless, from Figure \ref{fig:clustered_xrf} we may see that the relevant feature (i.e. the highest peaks) are related in the two pictures, while the gold dashed line presents a more noisy behaviour outside of relevant features, as expected (since there is no suppression of noise due to dimensionality of space in the latter case, and thus we have an apparent noise enhancement).

In the background of those plots, it is reported a flattened representation of the whole datacube, using the \textit{jet} colormap. On the Y-axis we see the pixel id (i.e. a linear combination of the $(i,j)$-pixel indices, namely $\mathrm{id(i,j)}= i + j\cdot \mathrm{width}$), while in the X-axis the Energy channel depth (which is shared with the aforementioned plots).

Figure \ref{fig:clustered_histograms} shows the three averages on the sample plot, and the background is the absolute difference of the flattened ``true" vs reconstructed data-cubes. 

Please notice that the Iterative K-Means in latent space has correctly - and independently - identified 4 clusters, i.e. the correct number of clusters identified by a similar, but completely unrelated, iterative clustering algorithm, performed in RGB space, which was thus used to generate the synthetic XRF. 

\begin{figure}[ht]
    \centering
    \includegraphics[width=0.95\textwidth]{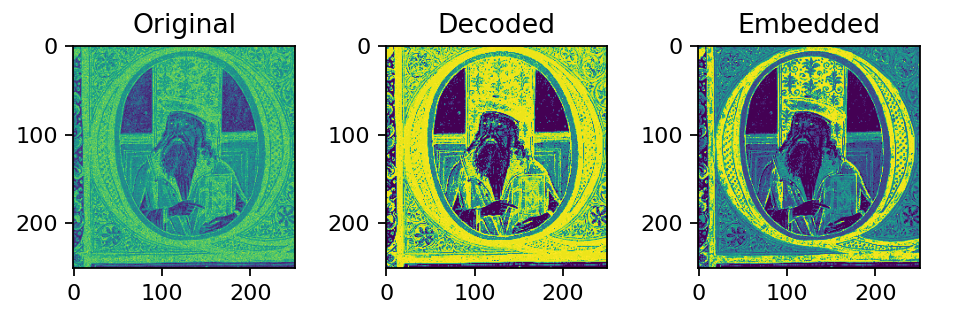}
    \caption{Energy integrated spectral data-cubes. }
    \label{fig:integrated_xrf}
\end{figure}

Similarly as in the previous case, we report in  Figure \ref{fig:integrated_xrf} the energy-integrated data-cube. It immediately shows an unexpected noise reduction (or signal enhancement), which adds readability to the image. Finally, the right image in Figure \ref{fig:integrated_xrf} represents the embedded energy-integrated datacube $E_{i,j} = \sum_n \enc[I_{i,j ; n}]]$, which further refine the readability. It is expected, since the non-linear low-dimensional embedding should project away less relevant directions, especially noise. 

\begin{figure}[t]
    \centering
    \includegraphics[width=0.96\textwidth]{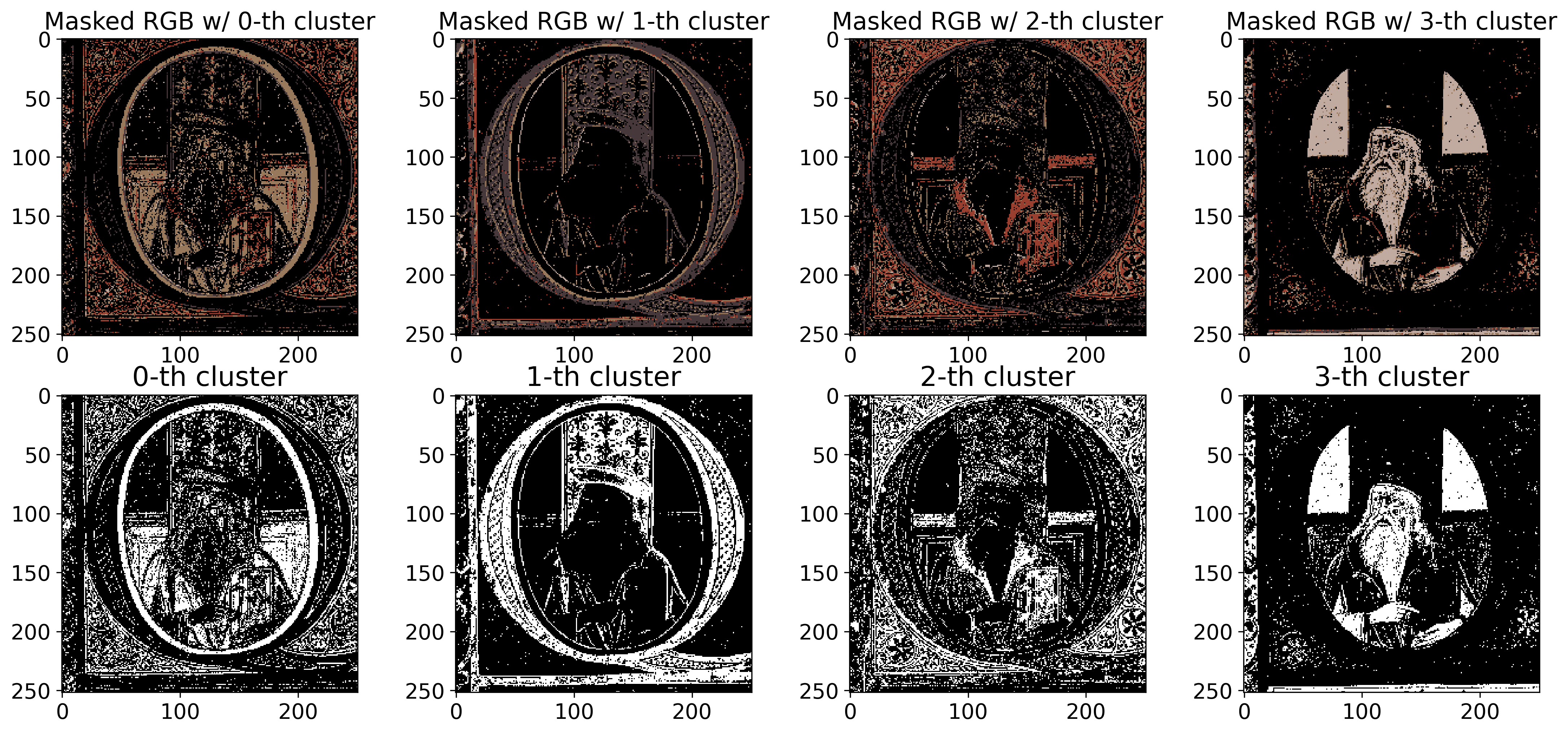}
    \caption{Masked clustered RGB image using clusters binary maps as masks.}
    \label{fig:masked_clustered_rgb}
\end{figure}

The last plot we present is Figure \ref{fig:masked_clustered_rgb}, where we report, in the first row, the masked clustered RGB appearing in Figure \ref{fig:1648_clustered_rgb}. The masking is done such that we have a black pixel, if that pixel does not belong to the cluster, and a coloured pixel, if the pixel belongs to the cluster. For example, in the 1-th cluster, it emerges the structure of the illuminated Q letter, which corresponds to the blueish cluster of Figure \ref{fig:1648_clustered_rgb}.
Thus, it emerges that each cluster has an average RGB colour which is similar to the one of the RGB cluster of Figure \ref{fig:1648_clustered_rgb}. 
The second row is just the cluster representation in black and white binary colour map.

\subsubsection{Check against RGB clustering:}\label{subsubsec:CH_check_vs_rgb}

We are now in the position to perform the same evaluation as the one done in Section \ref{subsubsec:Astro_check_vs_rgb};
we thus perform the same Iterative K-Means clustering approach to the RGB image, and use it as a baseline to compare it to the clustering obtained with the model for the Cultural Heritage dataset.

\begin{figure}[t]
    \centering
    \includegraphics[width=0.9\textwidth]{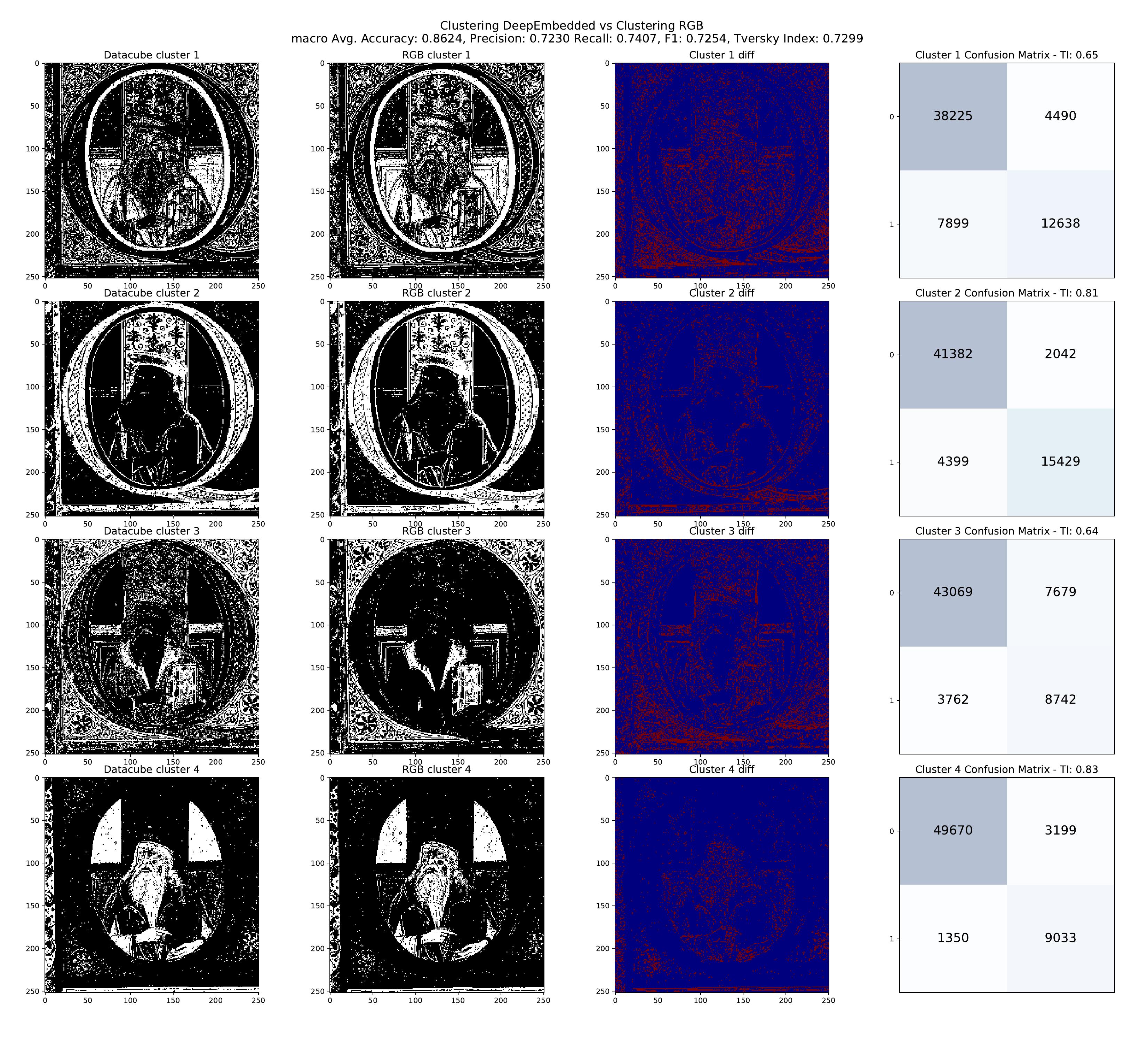}
    \caption{Deep Clustering Segmentation vs RGB Clustering for the Deep Variational Embedding model on Cultural Heritage synthetic datacube.}
    \label{fig:CH_DeepEmbedVsRGB_clustering}
\end{figure}

Again, in Figure \ref{fig:CH_DeepEmbedVsRGB_clustering} we report the result of the check, on a cluster-by-cluster basis. We recall that the columns are, from left to right: the black-and-white image of the cluster mask (i.e., pixels are white if they do belong to the cluster, and black if they don't) as obtained from the Deep Embedding Model; the same, but obtained from the RGB clustering; the image of mislabelled pixels (i.e., pixels appears red if they are mislabelled and blue if they are correctly labelled); the confusion matrix. On the confusion matrix is reported the number of counts (from left to right, from top to bottom: true positives, false positives, false negatives, true negatives). In the confusion matrix image subtitle is reported the computed Tversky Index.

\begin{figure}[t]
    \centering
    \includegraphics[width=0.6\textwidth]{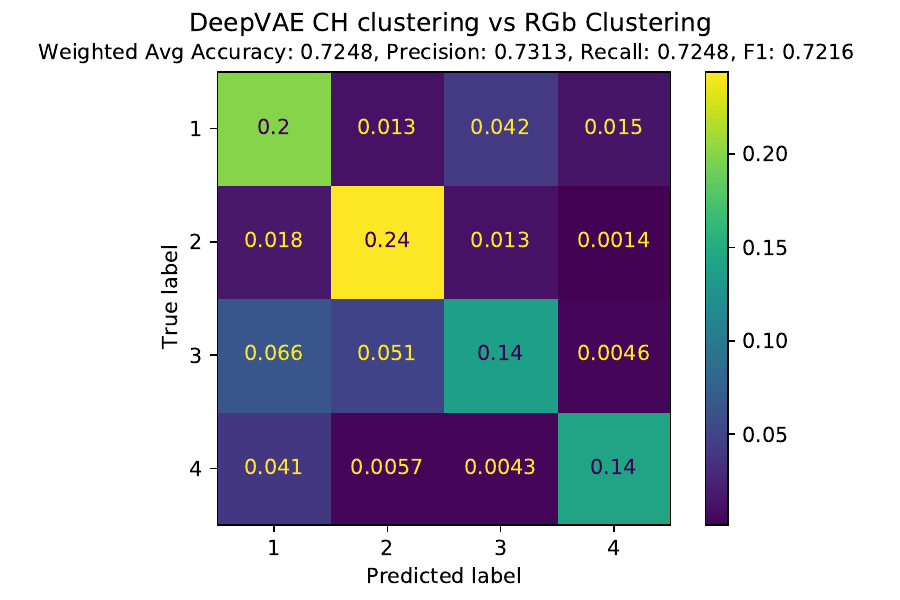}
    \caption{Multi-class Confusion Matrix of Deep Clustering Segmentation vs RGB Clustering on Cultural Heritage synthetic datacube.}
    \label{fig:CH_DeepEmbedVsRGB_clustering_mc_cm}
    
\end{figure}
In Figure \ref{fig:Astro_MC-ConfusionMatrix} we report the multi-class confusion matrix; on each element of the confusion matrix is reported the counts normalised to the whole population. In the Figure subtitle is reported the Weighted average of Accuracy, Recall, Precision, and F1 score. 


\section{Conclusions} \label{sec:conclusions}

In this contribution, we have demonstrated the feasibility of training a deep variational embedding model to perform spectral datacube segmentation through Deep Clustering in latent space. This approach leverages the power of Deep Learning to analyse and categorise (in a self-supervised manner) complex spectral datacubes, which are multi-dimensional arrays of data commonly used in various scientific fields.

We conducted our experiments using two models that were trained on two synthetic datasets. These models were designed to test the effectiveness of our approach in two distinct scenarios: an astrophysics case and a Heritage Science case.

In the astrophysics case, our deep embedding model was tasked with analysing and segmenting synthetic spectral datacubes that represent astronomical observations. These datacubes contain spectra of simulated ionised belonging to different classes, namely HII regions, shock regions and/or planetary nebulae, and our model was able to successfully segment this data, demonstrating its potential for use in astrophysical research.

In the Heritage Science case, our deep variational embedding model was used to analyse and segment synthetic spectral datacubes of MA-XRF imaging  on medieval Italian manuscripts. Such methods are often used in the study and preservation of Cultural Heritage artifacts (for example, but not limited to, pictorial artworks), and our model’s successful segmentation of this data shows its potential for aiding conservation scientists in their analysis.

In both cases, our deep variational embedding model was able to achieve successful segmentation of the spectral datacubes. This not only validates our approach but also opens up new possibilities for the application of Deep Learning techniques in the analysis of spectral datacubes in various scientific fields. Our findings pave the way for further research and development in this area, with the potential benefits either in the astrophysical and Heritage Science fields.


We are eagerly anticipating the application of this methodology to real-world data, encompassing both astrophysical and heritage scientific. This will be achieved by fine-tuning the models we have trained on synthetic data, adapting them to handle the complexities and nuances of real-world data.

Furthermore, we plan to use the cloud-based distributed computing infrastructure that is currently developed at Istituto Nazionale di Fisica Nucleare (INFN) and ICSC. Using software tools for distributed computing, such as Dask, will facilitate distributed inference, enabling us to rapidly compress spectral datacubes. This approach not only enhances the efficiency of our model but also significantly reduces the computational resources required, making the process more accessible and scalable.


\section{Code and Data Availability}\label{sec:data-and-code}

The code used in this project can be found at the ICSC Spoke 2 GitHub repository \cite{Bombini_FastExtendedVision_DeepCluster_GitHub_repository_2024}. 

The synthetic dataset \cite{dataset} and trained models \cite{alessandro_bombini_2024_143543} are available at the INFN Open Access Repository service. 

\section*{Acknowledgments}

\subsection*{Funding}
This work is supported by ICSC – Centro Nazionale di Ricerca in High Performance Computing, Big Data and Quantum Computing, funded by European Union – NextGenerationEU, by the European Commission within the Framework Programme Horizon 2020 with the project ``4CH - Competence Centre for the Conservation of Cultural Heritage"  (GA n.101004468 – 4CH) and by the project AIRES–CH - Artificial Intelligence for digital REStoration of Cultural Heritage jointly funded by the Tuscany Region (Progetto Giovani Sì) and INFN.

The work of AB was funded by Progetto ICSC - Spoke 2 - Codice CN00000013 - CUP I53C21000340006 - Missione 4 Istruzione e ricerca - Componente 2 Dalla ricerca all'impresa – Investimento 1.4.

The work of FGAB was funded by the research grant titled ``Artificial Intelligence and Big Data'' and funded by the AIRES-CH Project cod.~291514 (CUP I95F21001120008).


\clearpage
\appendix

\section{Datacube segmentation using Iterative K-Means Clustering after non-linear dimensional reduction with UMAP}\label{app:umap}

In this appendix we report a single-datacube application of the alternative pipeline described in \ref{subsubsec:DimRed}: first, we perform a non-linear dimensional reduction on the datacube spectra using the Uniform Manifold Approximation and Projection (UMAP) algorithm \cite{mcinnes2020umap}; then, we perform the iterative K-Means clustering to get an emergent datacube segmentation, as discussed for the Deep Embedding models in Section \ref{sec:methods}.

\subsection{Astrophysical Datacube}
We perform the same check done in Section \ref{subsubsec:Astro_check_vs_rgb} for the Deep embedding model. Using the standard UMAP code implementation of \cite{mcinnes2020umap}.

\begin{figure}[t]
    \centering
    \includegraphics[width=0.9\textwidth]{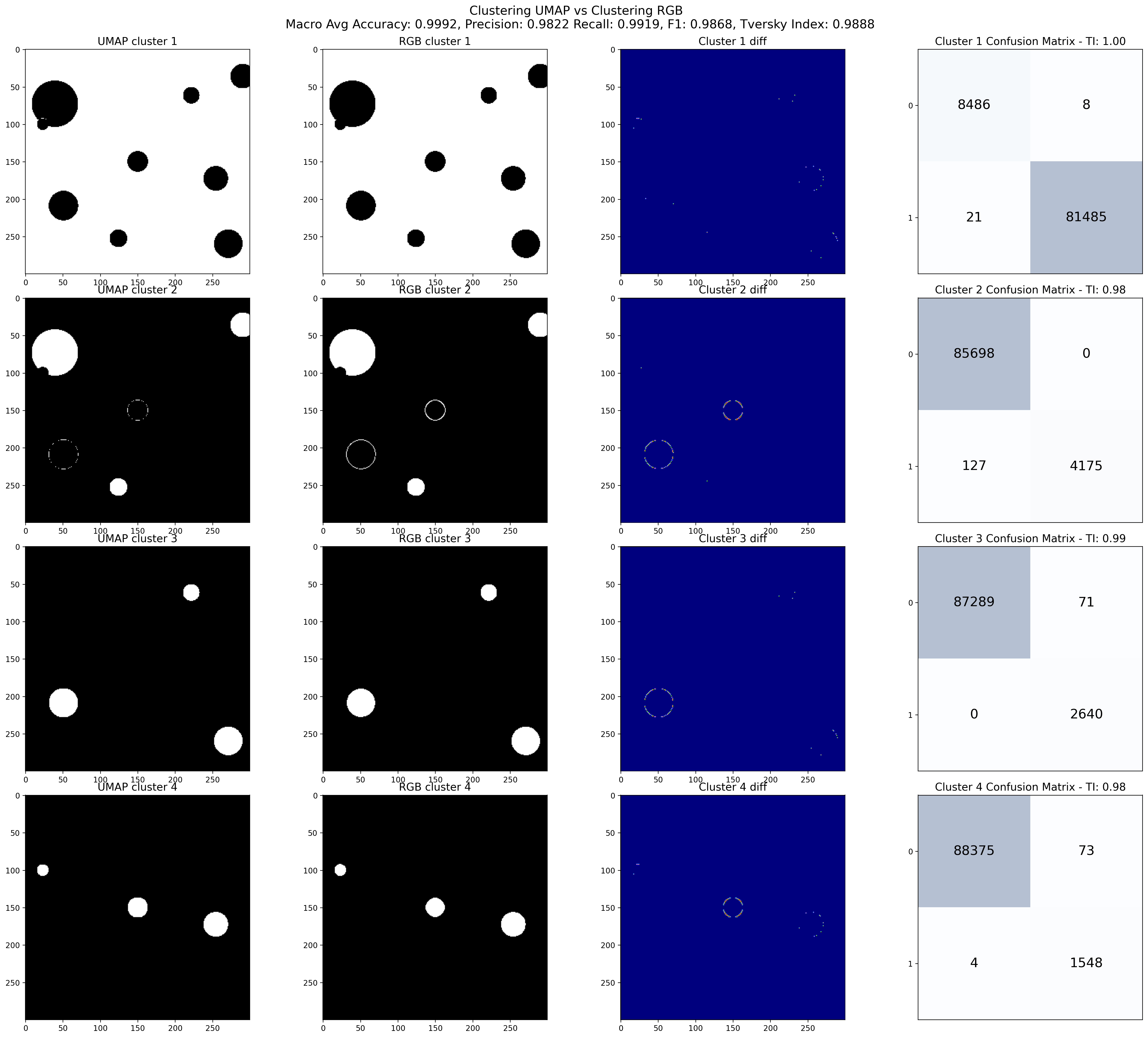}
    \caption{UMAP+Clustering Segmentation vs RGB Clustering Deep  Embedding model on Astrophysical synthetic datacube.}
    \label{fig:Astro_UMAPVsRGB_clustering}
\end{figure}

\begin{figure}[t]
    \centering
    \includegraphics[width=0.6\textwidth]{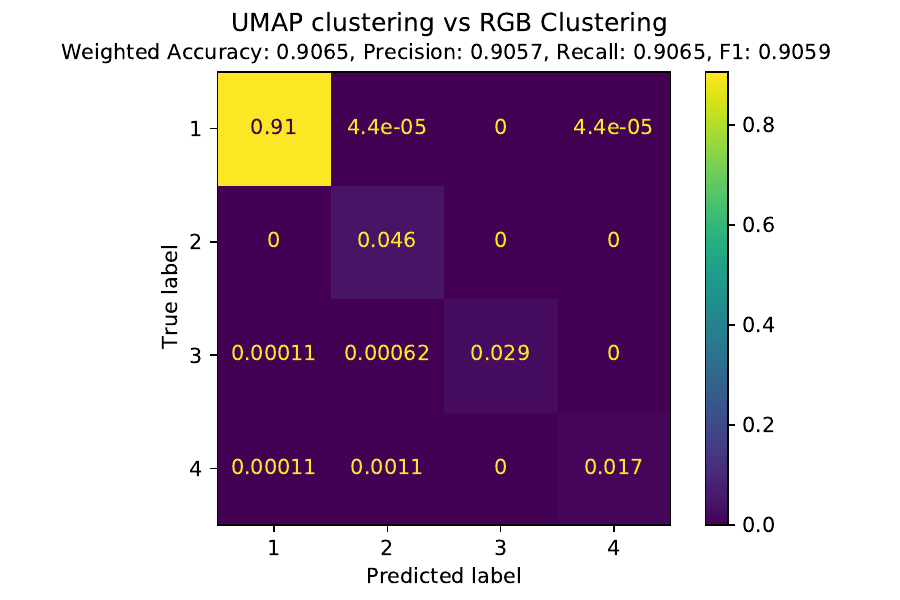}
    \caption{Multi-class Confusion Matrix of UMAP Segmentation vs RGB Clustering on the synthetic Astrophysical Datacube.}
    \label{fig:Astro_UMAP_MC-ConfusionMatrix}
\end{figure}

The results of the clustering are reported in Figures \ref{fig:Astro_UMAPVsRGB_clustering} and \ref{fig:Astro_UMAP_MC-ConfusionMatrix}.  The results are overall less noisy and have higher scores with respect to the Deep Embedding model ones, reported in Figures \ref{fig:Astro_DeepEmbedVsRGB_clustering} and \ref{fig:Astro_MC-ConfusionMatrix}. Nevertheless, the UMAP algorithm was trained on the whole datacube image, and thus it was also informed with background signal. That was not the case for the Deep Embedding model, which was nevertheless capable of correctly clustering most of the background points. 


\subsection{Cultural Heritage Datacube}
We now perform the same check done in Section \ref{subsubsec:CH_check_vs_rgb}, but with UMAP as a dimensional reduction algorithm. A similar approach was hinted in \cite{Ruberto2023}, but not described in depth. 

\begin{figure}[t]
    \centering
    \includegraphics[width=0.9\textwidth]{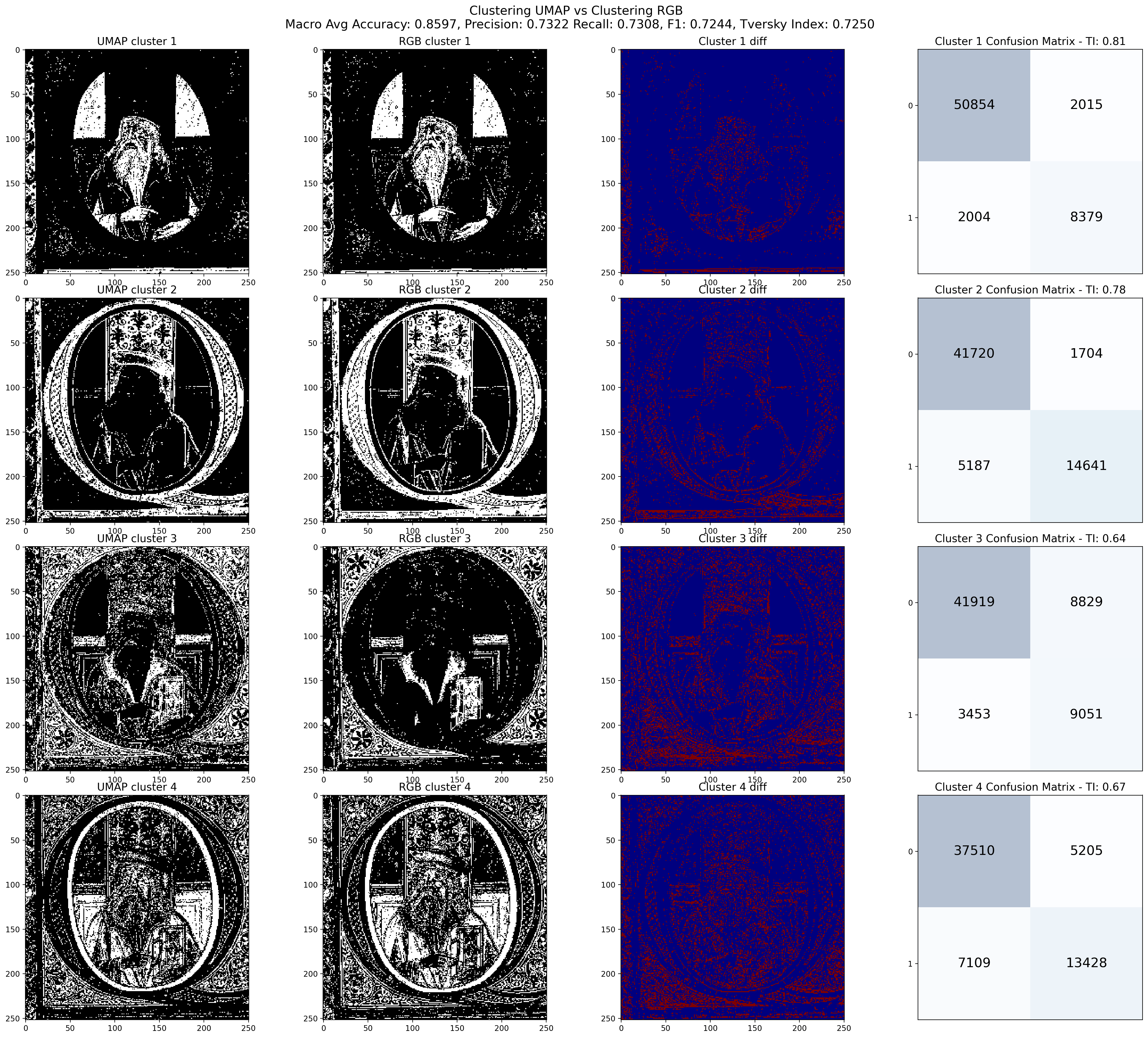}
    \caption{UMAP+Clustering Segmentation vs RGB Clustering Deep  Embedding model on CH synthetic datacube.}
    \label{fig:CH_UMAPVsRGB_clustering}
\end{figure}

\begin{figure}[t]
    \centering
    \includegraphics[width=0.6\textwidth]{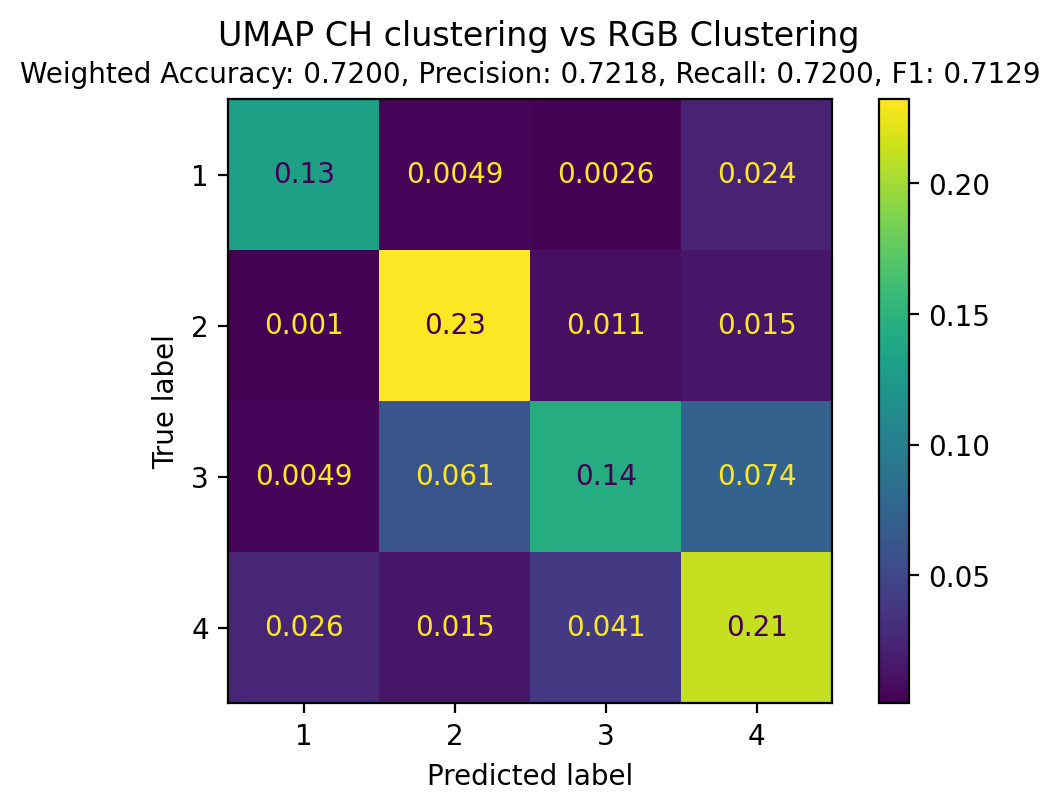}
    \caption{Multi-class Confusion Matrix of UMAP Segmentation vs RGB Clustering on the synthetic CH Datacube.}
    \label{fig:CH_UMAP_MC-ConfusionMatrix}
\end{figure}

The results of the clustering are reported in Figures \ref{fig:CH_UMAPVsRGB_clustering} and \ref{fig:CH_UMAP_MC-ConfusionMatrix}.
The results are in line with the one obtained in Section \ref{subsubsec:CH_check_vs_rgb}, even slightly worse;
this may due to the fact that this test case has a more complex spectral composition, as well as the fact that all the pixels signals are statistically similar to the ones used in the training phase, and there are no regions of out-of-distribution like in the previous case, where any background-like spectral signal has not been seen during training by the deep embedding model. 

Nevertheless, as described in Section \ref{subsubsec:DimRed}, while it is possible to obtain datacube segmentation with dimensional reduction algorithm followed by embedded space clustering, the features of the two approaches are  different, and are thus non-competing approaches, but rather two complementary ones. 

\section{Study of Model resilience against Signal Degradation} \label{app:noise}
In the astrophysics case, since the model used for generating the synthetic signal is capable of noise modelling, it allows us to perform an evaluation of the resilience of the trained model of Section \ref{subsec:Astro-Res} under the degradation of signal, by reducing the Peak-Signal-to-Noise ratio, without further re-training. This is due to the fact that autoencoders can also be used to remove noise (these methods are often called denoising autoencoders). \cite{10.1145/1390156.1390294}. 

In the following, we use the formulae
\begin{equation}
 \mathrm{PSNR}(X, Y) = 10 \log_{10} \frac{ \left(\max X \right)^2}{ \mathrm{mean} \left( X-Y \right)^2}    \,,
 \end{equation}
 \begin{equation}
 \mathrm{degradation}(X,Y) = 1 -  \frac{\mathrm{PSNR}(X, X) }{  \mathrm{PSNR}(X, Y)  } \,,
\end{equation}

where $X$ is the original signal, and $Y$ the one where we have added noise.

We thus re-perform, for each of the following degradation levels
\begin{equation}
    [0, 0.21, 0.33, 0.40, 0.45]
\end{equation}
the same analysis conducted in Section \ref{subsec:Astro-Res}. We report here the results of this analysis.

\begin{figure}[t]
    \centering
    \includegraphics[width=0.98\textwidth]{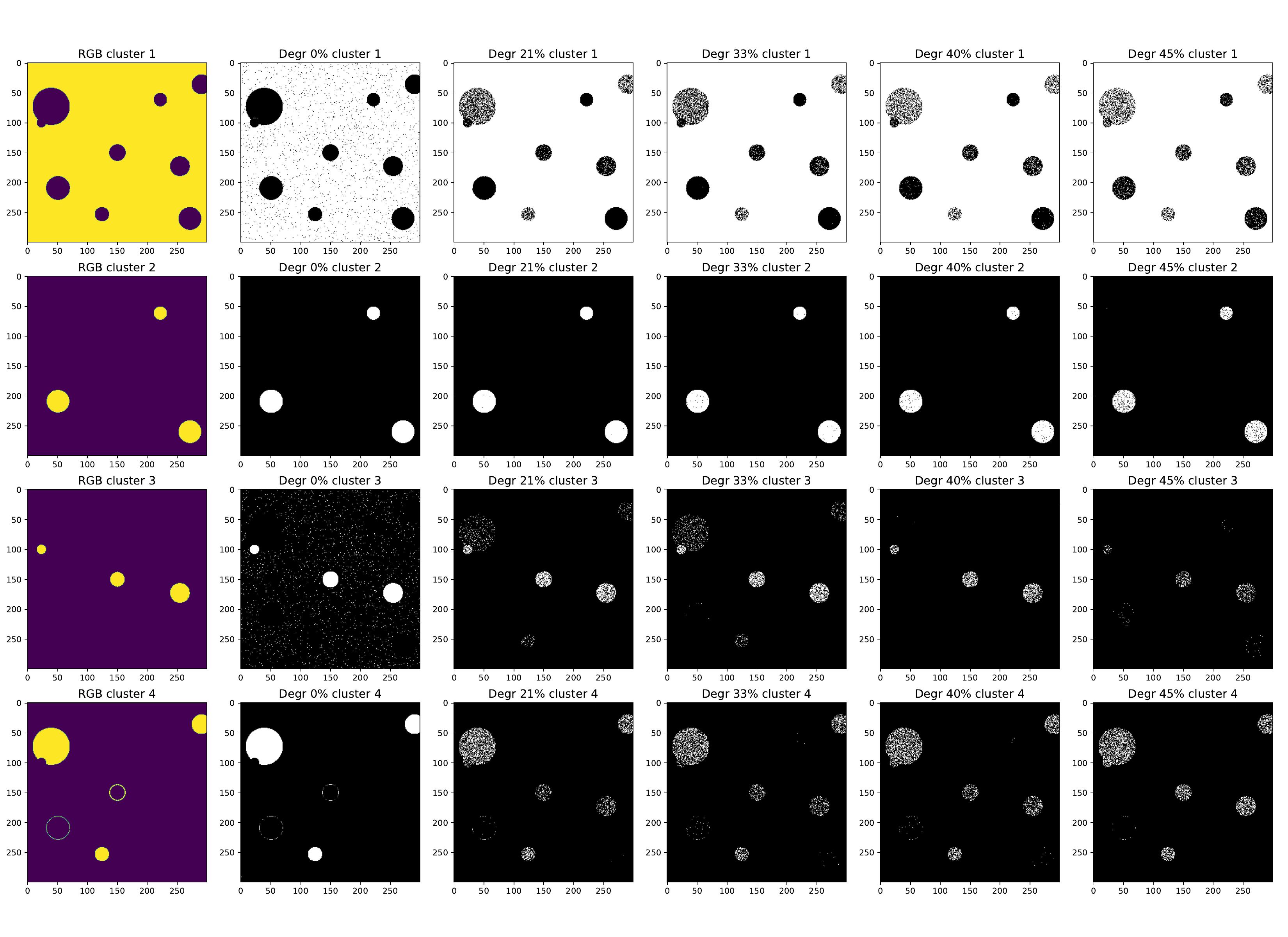}
    \caption{Effect of Noise addition in the Datacube Segmentation}
    \label{fig:noise_perc_overall}
\end{figure}

\begin{figure}[t]
    \centering
    \includegraphics[width = 0.8\textwidth]{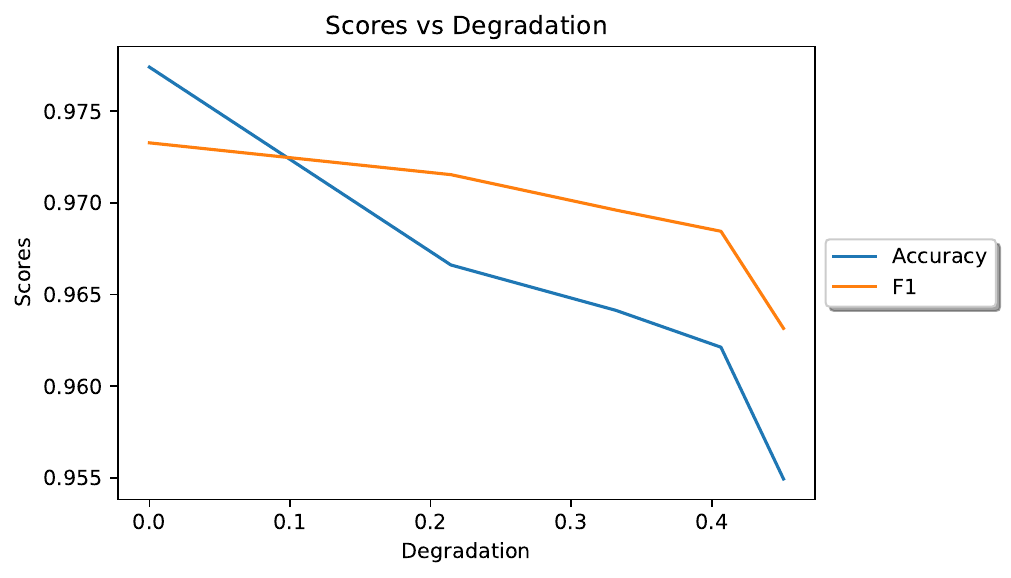}
    \caption{Loss of Weighted Accuracy and F1}
    \label{fig:noise_loss}
\end{figure}

In Figure \ref{fig:noise_perc_overall} we report the datacube segmentation as a function of the noise level; the rows represent the cluster, while the columns are defined as follows. The leftmost column, in \textit{viridis} colorscale, is the clustering obtained from the RGB image; then, from left to right, we show the corresponding found cluster at increasing values of noise. 

In Figure \ref{fig:noise_loss} we report how the noise levels affect the multi-class weighted accuracy and F1 score. 

We notice that the introduction of noise raises the misclassification of non-background signals, as expected.

\clearpage
\section*{References}
\bibliographystyle{iopart-num}  
\bibliography{references}

\end{document}